\pdfoutput=1
\documentclass[utf8]{article}

\usepackage[final,nonatbib]{neurips_data_2022}

\usepackage[T1]{fontenc}    % use 8-bit T1 fonts
\usepackage{url}            % simple URL typesetting
\usepackage{booktabs}       % professional-quality tables
\usepackage{amsfonts}       % blackboard math symbols
\usepackage{nicefrac}       % compact symbols for 1/2, etc.
\usepackage{microtype}      % microtypography
\usepackage{hyperref}
\usepackage{xcolor}         % colors

\usepackage[final,defaultcolor=orange,markup=underlined]{changes}

\usepackage{CJKutf8}        % Chinese characters
\usepackage{amsmath,amsfonts,bm}  % math
\usepackage{pifont}         % symbol
\usepackage{graphicx}       % graphics
\usepackage{diagbox}        % tables head
\usepackage{multirow}       % table
\usepackage{enumitem}       % list
\usepackage{siunitx}        % number
\usepackage[normalem]{ulem}
\usepackage{wrapfig}
\usepackage{tabularx}
\usepackage{subcaption}
\usepackage{tabu}

\usepackage{caption}
\def \vx{{\bm{x}}}
\def \vz{{\bm{z}}}
\def \xi{\vx^I}
\def \xt{\vx^T}
\def \zi{\vz^I}
\def \zt{\vz^T}

\hypersetup{
  colorlinks,
  linkcolor={red!80!black},
  citecolor={blue!80!black},
  urlcolor={green!50!black}
}

\newcommand{\tfbestfirst}[1]{\textcolor{magenta}{\textbf{#1}}}
\newcommand{\tfbestsecond}[1]{\textcolor{orange}{\underline{#1}}}
\newcommand{\tfbestthird}[1]{\textcolor{cyan}{\textit{#1}}}

\title{Wukong: A 100 Million Large-scale Chinese Cross-modal
  Pre-training Benchmark} %Dataset\\and A Foundation 

\author{
Jiaxi~Gu$^1$$^{*}$, Xiaojun~Meng$^1$$^{*}$, Guansong~Lu$^1$, Lu~Hou$^1$, Minzhe~Niu$^1$, Xiaodan~Liang$^2$$^\dagger$, \\ \textbf{Lewei~Yao$^1$, Runhui~Huang$^2$, Wei~Zhang$^1$, Xin~Jiang$^1$, Chunjing~Xu$^1$, Hang~Xu$^1$$^\dagger$}
}

\begin{document}
\begin{CJK*}{UTF8}{gbsn}

\maketitle

%% chapter: Abstract
\begin{abstract}
  Vision-Language Pre-training (VLP) models have shown remarkable
performance on various downstream tasks. Their success heavily relies
on the scale of pre-trained cross-modal datasets. However, the lack of
large-scale datasets and benchmarks in Chinese hinders the development
of Chinese VLP models and broader multilingual applications. In this
work, we release a large-scale Chinese cross-modal dataset named
Wukong, which contains 100 million Chinese image-text pairs collected from the web.
Wukong aims to benchmark different multi-modal pre-training methods to
facilitate the VLP research and community development.  Furthermore, we
release a group of models pre-trained with various image encoders
(ViT-B/ViT-L/SwinT) and also apply advanced pre-training techniques
into VLP such as locked-image text tuning, token-wise similarity in
contrastive learning, and reduced-token interaction. Extensive
experiments and a benchmarking of different downstream tasks including a new largest human-verified image-text test dataset are also provided. 
Experiments show that Wukong can serve as a promising
Chinese pre-training dataset and benchmark for different cross-modal
learning methods. For the zero-shot image classification task on 10
datasets, Wukong\textsubscript{ViT-L} achieves an average accuracy of
73.03\%.
For the image-text retrieval task,
it achieves a mean recall of 71.6\% on AIC-ICC which is 12.9\% higher than WenLan~2.0. Also, our Wukong models are benchmarked on downstream tasks with other variants on multiple datasets, e.g., Flickr8K-CN, Flickr-30K-CN, COCO-CN, et~al. More information can be referred to
\url{https://wukong-dataset.github.io/wukong-dataset/}.

  \let\thefootnote\relax\footnotetext{$^1$ Huawei Noah's Ark Lab \hspace{2mm}  $^*$ These two authors contribute equally.} 
  \let\thefootnote\relax\footnotetext{$^2$ Sun Yat-sen University \hspace{2mm} $^\dagger$ Corresponding authors: \texttt{xu.hang@huawei.com} \& \texttt{xdliang328@gmail.com}}

\end{abstract}

%% chapter: Introduction
\section{Introduction}

Pre-training large-scale models on big data, and fine-tuning them on
downstream tasks, \added{has}
% \deleted{have}
become an emerging paradigm of artificial
intelligence systems. Models such as BERT~\cite{devlin2018bert} and
GPT~\cite{brown2020language} grow in popularity in the natural
language processing community as they possess high transferability to
a wide range of downstream tasks,
% or even zero-shot tasks\footnote{\hou{not quite consistent with the pretrain and "finetune" paradigm}}
yielding
state-of-the-art performance. Recent works such as
CLIP~\cite{radford2021learning}, ALIGN~\cite{jia2021scaling}, and
FILIP~\cite{yao2021filip} further extend this paradigm to the joint
Vision Language Pre-training (VLP) domain and show superior results
over state-of-the-art methods on various downstream tasks. \added{Meanwhile, VLP models can be easily adapted to multiple practical applications such as image search engines, multi-choice visual answering and image labelling. In general,} this
promising direction draws significant attention from both industry and academia
to consider it as the path to the next-generation AI models.

Two reasons lead to the success of VLP models. On the
one hand, more advanced model architectures such as
ViT~\cite{dosovitskiy2020image}/BERT~\cite{devlin2018bert} and
training objectives like contrastive learning~\cite{he2020momentum},
are usually able to lift the powerful generalization and robustness
capabilities of learned representations. On the other hand, thanks to
the concurrent advancement in
hardware~\cite{stuart2011multi,kindratenko2009gpu} and distributed
training
frameworks~\cite{narayanan2021efficient,rajbhandari2020zero,rasley2020deepspeed},
more and more data can be fed into a large-scale model to improve the
generalization, transferability and zero-shot capability. In either
vision or language tasks, pre-training on larger-scale data such as
JFT-300M~\cite{sun2017jft300m} in image classification
\cite{riquelme2021scaling}, C4 dataset in
T5~\cite{raffel2020exploring}, has been proven useful and critical
for improving downstream task performance via transfer or prompt
learning. In addition, recent work~\cite{jia2021scaling} has already
shown the potential of scaling up the VLP model by more than 100
million noisy image-text pairs from the web.

\begin{wraptable}{l}{7.4cm}
\vspace{-2mm}
  \centering
  \setlength{\tabcolsep}{1pt}
  \caption{\label{tab:public-datasets} An overview of VLP datasets.}
  \footnotesize  % use small font size in this table
  \begin{tabular}{r|ccr}
    \textbf{\shortstack{Dataset}} & \textbf{\shortstack{Language}} & \textbf{\shortstack{Avail\\-ability}} & \textbf{\shortstack{Image-text\\pairs}}\\
    \midrule
    Flickr30k~\cite{young2014flickr30k} & English & \ding{51} & \num{31783}\\
    CxC~\cite{parekh2020cxc} & English & \ding{51} & \num{247315}\\
    SBU Captions~\cite{ordonez2011sbucaptions} & English & \ding{51} & \num{1000000}\\
    Product1M~\cite{capture} & Chinese & \ding{51} & \num{1000000}\\
    CC12M~\cite{changpinyo2021cc12m} & English & \ding{51} & \num{12000000}\\
    RedCaps~\cite{desai2021redcaps} & English & \ding{51} & \num{12011111}\\
    YFCC100M~\cite{thomee2016yfcc100m} & English & \ding{51} & \num{99200000}\\
    WIT~\cite{srinivasan2021wit} & multilingual & \ding{51} & \num{11500000}\\
    LAION-400M~\cite{schuhmann2021laion400m} & English & \ding{51} & \num{400000000}\\
    \midrule
    JFT-300M~\cite{sun2017jft300m} & English & \ding{55} & \num{300000000}\\
    JFT-3B~\cite{zhai2021jft3b} & English & \ding{55} & \num{3000000000}\\
    IG-3.5B-17k~\cite{mahajan2018ig35b} & English & \ding{55} & \num{3500000000}\\
    M6-Corpus~\cite{lin2021m6} & Chinese & \ding{55} & \num{60500000}\\
    \midrule
    \textbf{Wukong} & \textbf{Chinese} & \ding{51} & \textbf{\num{101483885}}\\
    \bottomrule
  \end{tabular}
  \vspace{-2mm}
\end{wraptable}

Therefore, the success of VLP models pre-trained on large-scale data
urges people to continuously crawl and collect larger image-text
datasets.
Table~\ref{tab:public-datasets} shows an overview of many
popular datasets in the VLP domain.
For English datasets, the publicly available
Flickr30k~\cite{plummer2015flickr30k}, SBU
Captions~\cite{ordonez2011im2text}, and
CC12M~\cite{sharma2018conceptual} are relatively small,
while LAION-400M \cite{schuhmann2021laion400m} is several magnitudes larger.
Despite the availability of large-scale English datasets, directly translating them into Chinese and then training a Chinese VLP model can lead to a severe performance drop.
We speculate this is due to the existence of many
Chinese idioms and slang that simple translation cannot cover but brings errors that harm the performance. The current community lacks a large-scale publicly
available dataset in Chinese, resulting in (a) the development of the community being stunted; (b) secret large datasets used to achieve surprisingly good performance that other works cannot fairly compare with.

\begin{wraptable}{r}{5.0cm}
\vspace{-2mm}
  \centering
  \setlength{\tabcolsep}{2pt}
  \caption{\label{tab:test-datasets} Comparison of multimodal Chinese retrieval benchmarks.}
  \footnotesize  % use small font size in this table
  \begin{tabular}{c|rr}
    \textbf{Dataset} & \textbf{\#Images} & \textbf{\#Texts}\\
    \midrule
    Flickr8K-CN\textsubscript{Test} & \num{1000} & \num{5000}\\
    Flickr30K-CN\textsubscript{Test} & \num{1000} & \num{5000}\\
    COCO-CN\textsubscript{Test} & \num{1000} & \num{1053}\\
    AIC-ICC\textsubscript{Test-1} & \num{30000} & \num{150000}\\
    AIC-ICC\textsubscript{Test-2} & \num{30000} & \num{150000}\\
    MUGE\textsubscript{Test} & \num{30399} & \num{5004}\\
    \midrule
    \textbf{Wukong-Test} & \textbf{\num{33365}} & \textbf{\num{33365}}\\
    \bottomrule
  \end{tabular}
  \vspace{-2mm}
\end{wraptable}

To bridge this gap, we release a large-scale Chinese cross-modal
dataset named Wukong, which contains
100 million image-text pairs collected from the
web. To guarantee the diversity and generalization, our Wukong dataset is
collected according to a high-frequency Chinese word list with 200K
queries. We also adopt image-based and text-based filtering strategies
for further refinement. The resulting dataset is currently the largest
Chinese vision-language dataset.
% so far. 
We perform an analysis of this
dataset and show that it covers a wide range of visual and textual
concepts. Besides, we also build a test set
called \emph{Wukong-Test}, the quality of which has been verified by human experts.
\added{From the feedback, the image-text consistency is guaranteed in general even
if all the data are collected on the web and only some simple filtering strategies
are applied. Specifically, there are only about 2\% image-text pairs are marked as weakly corresponding.} Table~\ref{tab:test-datasets} shows the
comparison of available Chinese image-text testing datasets.

Training a large-scale VLP model is quite expensive. For example, the
largest CLIP~\cite{radford2021learning} model takes 18 days to train
on 592 NVIDIA-V100
% V100 
GPUs and M6-10T~\cite{lin2021m6} is trained on 512
NVIDIA-V100 
GPUs for around 10 days. Thus it is almost impossible for
everyone to pre-train a large-scale model due to substantial financial
costs and hardware requirements. It is in great demand for researchers
to download and reuse various kinds of pre-trained large-scale
% \footnote{\hou{
Chinese
% ? There are already many English VLP models online.}}
VLP
models. However, the choices of publicly available large VLP models
are also very limited, which hinders the improvement of performance on
downstream tasks of large-scale models.

To contribute to the community, we release a group of dual-stream VLP models pre-trained using different image encoders
(ViT~\cite{dosovitskiy2020image} and SwinT~\cite{liu2021swin})
and different pretraining techniques (CLIP~\cite{radford2021learning},
FILIP~\cite{yao2021filip}, and LiT~\cite{zhai2021lit}).
We further provide an extensive Chinese
benchmarking on various downstream tasks and datasets with hand-crafted Chinese labels, such as zero-shot image classification and image-text retrieval. 
Interestingly, though the frozen image encoders are trained on English image-text pairs,  directly aligning them with a trainable Chinese text encoder still achieves remarkable performance on downstream tasks. This also indicates the strong cross-lingual generalization of these pre-trained image encoders. 
Besides, we also find that using the cross-modal token-wise similarity from
FILIP maintains the fine-grained word-patch alignment for various image encoders, even when they are frozen during the contrastive learning. Moreover, compared with the Chinese word-grained tokenization, we find that using character-grained tokenization in our models achieves better performance. More findings can be found in Section~\ref{sec:expt}.

Experiments show that Wukong can serve as a promising Chinese
pre-training dataset for different cross-modal learning methods. The
pre-trained models show prominent performance on various downstream
tasks such as zero-shot image classification and image-text
retrieval. Specifically, our model Wukong\textsubscript{ViT-L}, pre-trained using Wukong dataset, achieves up to 73.03\% average top-1 accuracy on 10 datasets for zero-shot image classification. It also achieves 71.6\% mean recall on AIC-ICC for image-text retrieval. This result is higher than that of WenLan~2.0, which is a Chinese image-text multimodal model pre-trained on its own large-scale dataset, by 12.9\%.

% Specifically, for zero-shot image classification, Wukong\textsubscript{ViT-L}
% reaches up to 73.03\% average top-1 accuracy on 10 datasets. For the image-text
% retrieval task, our best model significantly outperforms WenLan~2.0 on
% AIC-ICC by 10\% of top-1 recall for image-to-text retrieval, and
% 11.6\% of top-1 recall for text-to-image retrieval
% respectively. Visualization on word-patch alignment also shows that
% our model learns meaningful finer-grained features via the token-wise
% similarity.

In summary, our main contributions are:
% \vspace{-2mm}
\begin{itemize}[noitemsep, nolistsep, leftmargin=*]
%\begin{enumerate}[label=(\alph*),leftmargin=*,topsep=1pt]
\item We release a large-scale Chinese VLP dataset with 100 million image-text pairs, covering a wide range of concepts. We also provide various benchmarking datasets with human-verified image-text pairs and Chinese labels for benchmarking the performance.
\item We release a group of large-scale VLP models pre-trained with
  various popular architectures and methods. An extensive study and benchmarking
  are also provided.
\item Our pre-trained model shows state-of-the-art performance on
  Chinese benchmarks such as zero-shot image classification and image-text retrieval tasks.
%\end{enumerate}
\end{itemize}

%% chapter: Related work
\vspace{-2mm}
\section{Related Work}
\vspace{-2mm}
\textbf{Vision-Language Pre-training (VLP) Models.} There are two typical architectures of VLP models according to the
modality interaction methods, i.e., single-stream and
dual-stream. Single-stream models~\cite{kim2021vilt,li2019visualbert}
directly concatenate the visual and textual embeddings together and
feed them to a single transformer-based model. This kind of model 
can
be easily fit into
text/image generation tasks to perform image
captioning or text-to-image generation, which are usually hard to
evaluate and benchmark. Dual-stream models such as
ViLBERT~\cite{lu2019vilbert}, CLIP~\cite{radford2021learning}, and
ALIGN~\cite{jia2021scaling} have separate models for each
modality. This paradigm is more flexible and efficient when modeling
each modality, e.g., CNN for images and Transformers for texts. Moreover,
dual-stream models have the merit of efficient inference for downstream
tasks such as image-text retrieval, since the two encoders can be decoupled and the image/text features can be pre-computed offline.
In
CLIP~\cite{radford2021learning}, the authors also evaluate the image
encoder as a self-supervised pre-trained model and show promising
results. This paper mainly follows and benchmarks the dual-stream
approaches.

\textbf{Vision-Language Datasets.} The current success of VLP models greatly lies in the scale of
pre-trained datasets. The publicly available pre-training datasets
used by recent VLP models are mainly image caption data or image-text
pair data. Many small-sized datasets (e.g., a few hundred thousand)
such as COCO-Captions~\cite{lin2014microsoft},
Flickr30k~\cite{plummer2015flickr30k}, Visual
Genome~\cite{krishna2016visual}, and VQA2~\cite{goyal2017making} are
hand-annotated data that have very limited domain and diversity. On
the other hand, pre-training models on online collected data (such as
alt-texts from the HTML pages) have shown promising
results. CC3M~\cite{sharma2018conceptual},
CC12M~\cite{changpinyo2021cc12m} and
YFCC100M~\cite{thomee2016yfcc100m} have millions of image-text pairs
in English generated by an online data collection pipeline including
image and text filters, as well as text transformations. VLP models on
these datasets have shown to be effective in multiple downstream
tasks. Moreover, larger-scale datasets with more than 100M samples
(e.g., CLIP~\cite{radford2021learning}: 400M and
ALIGN~\cite{jia2021scaling}): 1.8B) have even armed the recent VLP
models with surprisingly good zero-shot recognition ability, but they
are not publicly available. \added{In terms of vision-language datasets specifically for Chinese, as shown in Table~\ref{tab:public-datasets}, the dataset is either small-scale (Product1M~\cite{capture}) or private (M6-Corpus~\cite{lin2021m6}).} Thus, the current community lacks a
large-scale Vision-Language dataset in Chinese. We aim to contribute a
 Chinese dataset to benchmark various VLP methods.
%% Xuhang

%% chapter: Dataset collection (ref. ONCE, SODA) // Guansong, Minzhe
\vspace{-2mm}
\section{Construction of Wukong Dataset}
\vspace{-2mm}
% \begin{figure}[!t]
%   \centering \includegraphics[width=.95\linewidth]{figures/samples}
%   \caption{Examples of image-text pairs in our Wukong dataset. A 
%   diverse range of concepts are included.}
%   \label{fig:dataset_samples}
% \end{figure}

In this paper, we construct a dataset called Wukong containing 100 million
image-text pairs collected from the web. To cover as diverse 
concepts as possible, a series of keywords are taken as the starting point.
The original keyword list is taken from~\cite{song2018directional} and 
only the first \num{200000} most frequently seen keywords are used. These keywords are then used to search for images and their corresponding captions in Baidu,
a commonly used search engine for Chinese. For data balance, at most 
1000 image-text pairs are kept for each keyword.
% After the query list is constructed, we send each
% query to Baidu Image Search Engine, to get a list of image links and
% corresponding caption information. To keep a balance between different
% queries, we search for at most 1000 samples per query. Images are then
% downloaded with previously-obtained image links.
In this way, we
collect a total of 166 million raw \textlangle{}image,
text\textrangle{} pairs. Then, following common
practices~\cite{sharma2018conceptual,changpinyo2021cc12m,jia2021scaling},
we apply a series of filtering strategies described in the sections below to finalize Wukong dataset. Some examples in our dataset can be found in the appendix.
We also provide various benchmarking datasets with human-verified image-text pairs and Chinese labels for model benchmarks.
% the performance. 
Wukong-Test dataset contains 33k human-verified image-text pairs, which is currently the largest multimodal Chinese retrieval benchmark.

\textbf{Image-based Filtering.} We first filter the data according to the size and aspect ratio of the
image. Only images with both dimensions greater than 200 pixels, and
the ratio of large-to-small dimension \added{of at most 3}
% \deleted{is no more than 3}
are kept. In this way, we filter out images that are too small, too
tall or too wide. This kind of image is of poor quality, especially after data augmentation processes such as upsampling or square cropping.

\begin{wraptable}{r}{7.4cm}
  \centering
    \renewcommand{\arraystretch}{0.8}
  \vspace{-2mm}
  \setlength{\tabcolsep}{2pt}
\caption{\label{tab:wukong_dataset_statistics} Statistics of datasets.}
  \footnotesize  % use small font size in this table
\begin{tabular}{lrrccc}
  \toprule
  & \multirow{2.5}{*}{\shortstack{Image-text\\Pairs}} & \multirow{2.5}{*}{\shortstack{Unique\\Tokens}} & \multicolumn{3}{c}{\small{Tokens per Caption}} \\
  \cmidrule{4-6}
   & & & \small{mean} & \small{std} & \small{median} \\
  \midrule
  \small{Wukong} & \small{\num{101483885}} & \small{\num{20442}} & \small{\num{22}} & \small{\num{7}} & \small{\num{24}} \\
  \small{Wukong-Test} & \small{\num{33365}} & \small{\num{5155}} & \small{\num{22}} & \small{\num{7}} & \small{\num{24}} \\
  \bottomrule
\end{tabular}
\vspace{-2mm}
\end{wraptable}

\textbf{Text-based Filtering.} Secondly, to select samples with high-quality Chinese descriptions of
the corresponding image, we filter the data according to language,
text length, and the frequency of text accompanying an image. Specifically,
we first check the language and text length. We keep sentences that contain
at least one but fewer than 32 Chinese characters. We also discard
meaningless image descriptions like ``000.jpg'' from the text.
Texts paired with too many images are usually irrelevant to
the content of the images, like ``\texttt{查看源网页}'' (\textit{View
  source page}), ``\texttt{展开全文}'' (\textit{Expand text}),
``\texttt{摄影部落}'' (\textit{Photography community}). In practice,
we set this threshold as 10, i.e., we discard the image-text pairs
whose text appears more than 10 times in the whole corpus
collected. To protect the privacy of the individuals appearing in the
text, we substitute person names with a special token ``\textlangle{}人
名\textrangle{}'' (\textlangle{}\textit{Person
  name}\textrangle{}). Besides, we also construct a list of Chinese
sensitive words, and image-text pairs containing sensitive words are
also discarded.

After applying the above filtering strategies, we finally get a
dataset
% \deleted{of about 100 million \textlangle{}image, text\textrangle{} pairs}
\added{called Wukong for pre-training and a dataset called Wukong-Test for model testing. Table~\ref{tab:wukong_dataset_statistics} shows the statistics of them.}
% \deleted{As illustrated in Table~\ref{tab:wukong_dataset_statistics}, there are \num{20442} unique tokens within the texts of Wukong and the average number of tokens in each caption is 22.}

% Additionally, in Figure~\ref{fig:word-cloud}, we visualize the
% distribution of words (consisting of one or more tokens) in our
% dataset. We use the Chinese text segmentation module
% \textnormal{jieba}\footnote{\url{https://github.com/fxsjy/jieba}} to
% generate words and build this word cloud of our dataset.

% \begin{figure}[!t]
%   \centering
%   \includegraphics[width=.8\linewidth]{figures/word-cloud.png}
%   \caption{\label{fig:word-cloud} The word cloud generated with texts
%     in Wukong dataset. For example, ``\texttt{月}'' means
%     \textit{month}; ``\texttt{日}'' is \textit{day}; ``\texttt{做}''
%     is \textit{do} and ``\texttt{一个}'' means \textit{one}.}
% \end{figure}

%% chapter: Methodology
\vspace{-2mm}
\section{Methodology}
\vspace{-2mm}
\subsection{Text-Image Joint Alignment}
\vspace{-2mm}

\begin{figure}
\vspace{-2mm}
  \centering \includegraphics[width=.98\linewidth]{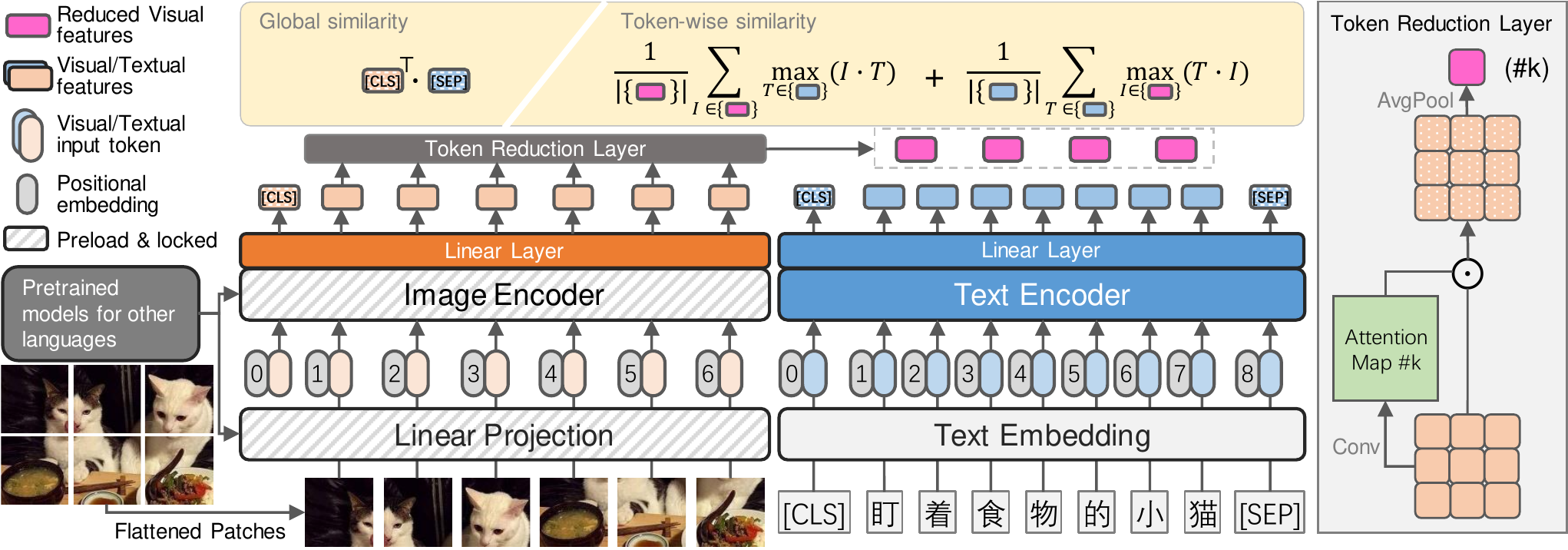}
  \caption{\label{fig:model-arch} Overviews of our released models. Our Chinese pre-trained models consist of an image encoder and a text encoder with visual tokens and textual tokens
    as inputs. We have three variations of pretrained models: global similarity (\textbf{CLIP}-style); token-wise similarity (\textbf{FILIP}-style) and token-wise similarity with token reduction layer (\textbf{Wukong}-style).}
\vspace{-2mm}
\end{figure}

Following the recent widely adopted
contrastive pre-training architectures~\cite{radford2021learning,yao2021filip}, we use a dual-stream model with
Transformer-based text and image encoders as shown in Figure~\ref{fig:model-arch}. These two encoders convert
textual and visual input tokens to embeddings of the same dimension.
In this learned joint embedding space, we use a contrastive loss to
encourage the paired image and text to have similar embeddings, while
non-paired ones to have distinct embeddings.

% Xiaohua Zhai et~al.~\cite{zhai2021lit} proposes that a locked pre-trained image encoder with an unlocked text encoder works well in contrastive learning. Inspired by their work, 
% we extend this idea to cross-lingual
% data sources, and try to align a locked image encoder pre-trained on
% English data sources with a trainable Chinese text encoder.
% The experiment results shown in Section~\ref{sec:ablation}
% confirm the effectiveness of our method.

% In~\cite{radford2021learning} and~\cite{jia2021scaling}, this cross-modal
% similarity is computed via dot product between the global features of
% the entire image and that of the text sequence. On the other hand,
% in~\cite{yao2021filip}, the similarity is computed based on a
% finer-grained interaction between the image patches and textual
% tokens, which also brings surprisingly good word-patch alignment and
% learns meaningful fine-grained features with promising localization
% ability.

\vspace{-2mm}
\subsection{Model Architectures}
\vspace{-2mm}
\textbf{Visual Encoder.} Two types of visual encoders,
\textit{i.e.,} Vision Transformer~\cite{dosovitskiy2020image} (ViT) and Swin
Transformer~\cite{liu2021swin} (SwinT), are used as backbones for training different model variants. For
ViT, the input image is first rescaled into a standard size and then
split into fixed-size patches. Each patch is linearly
embedded via a trainable linear projection. The resulting sequence of patch vectors is fed
to a standard transformer encoder. Different from ViT, SwinT uses
a hierarchical transformer that computes representation with shifted
windows, which accelerates the original self-attention computation to
non-overlapping local windows while also allowing for cross-window connection.

\textbf{Textual Encoder.} The textual encoder is a standard decoder-only
transformer as in~\cite{radford2021learning}. We use
WordPiece~\cite{wu2016wordpiece} with a vocabulary size of
\num{21128} for Chinese text tokenization. Similar
to~\cite{pires2019multilingual}, we add spaces around Chinese
characters before applying WordPiece so that Chinese is effectively
character-tokenized. We add two special tokens (i.e., 
\texttt{[CLS]} and \texttt{[SEP]}) at the beginning and ending 
of each text sequence. The text encoder has \num{12} layers,
each of which has \num{8} attention heads and a hidden state
dimension of \num{512}.

\textbf{Linear Projection of the Encoders.} On the top of the visual and textual encoders, the global representations of visual token sequence (e.g., \texttt{[CLS]} token for ViT; average pooled
representation of all patch tokens for Swin Transformer) and textual
token sequence (e.g., textual \texttt{[SEP]} token) are linearly
projected to the common multi-modal space, followed by
L2-normalization separately.

\textbf{Token Reduction Layer.} Instead of only computing the cross-modal similarity between global representations of sequences, we experiment
with a late interaction method as introduced in
FILIP~\cite{yao2021filip}. We aim to take into account the
fine-grained token-wise interaction between image patches and text
tokens. It could
potentially mine more detailed semantic word-patch alignment between
two modalities. Meanwhile, as a large amount of computation is introduced by this token-wise interaction, we propose a token reduction layer inspired by \cite{ryoo2021tokenlearner}. It aims to learn a small set of tokens (e.g., 12 or 24) from the whole output tokens of the visual encoder (e.g., 16$\times$16 in ViT-L/14), and use them for the reduced-token interaction. This token reduction layer is used in all the Wukong-style models.
% as described in Section~\ref{tklearner}
% The token reduction layer can largely reduce the computational cost for the late interaction method.

\vspace{-2mm}
\subsection{Pre-training Objectives}\label{method:objective}
\vspace{-2mm}
Cross-modal contrastive learning, typically represented by CLIP~\cite{radford2021learning}, is one effective
approach for training models using paired image-text data. It can
learn representations of two modalities simultaneously by
distinguishing the paired and unpaired samples. 
% The two sets of image and text samples are denoted as $\mathcal{I}$ and $\mathcal{T}$ respectively.
Given an image sample $\xi \in \mathcal{I}$ and a text sample
$\xt \in \mathcal{T}$, the training objective is to make the
learned image and text representations in the joint multi-modal space
close if they are paired \added{and far otherwise.}
% \deleted{while apart otherwise}
For a training batch consisting of $b$
image-text pairs $\{\xi_k, \xt_k\}_{k=1}^b$, $\xt_k$ (resp. $\xi_k$)
is positive to $\xi_k$ (resp. $\xt_k$) while negative to all other
texts (resp. images) in the same batch. Therefore, the image-to-text
and text-to-image contrastive losses for ($\xi_k$, $\xt_k$) can be
formulated as
\begin{math}
  \mathcal{L}^I_k(\xi_k, \{\xt_j\}_{j=1}^b) = -\frac{1}{b}\log\frac{\exp(s^I_{k,k})}{\Sigma^b_{j=1} \exp(s^I_{k,j})}
\end{math}
and
\begin{math}
\mathcal{L}^T_k(\xt_k, \{\xi_j\}_{j=1}^b) = -\frac{1}{b}\log\frac{\exp(s^T_{k,k})}{\Sigma^b_{j=1} \exp(s^T_{k,j})}
\end{math}
where $s^I_{k,j}$ denotes the similarity of the $k$-th image to the
$j$-th text, while $s^T_{k,j}$ denotes the similarity between the
$k$-th text to the $j$-th image. The total loss $\mathcal{L}$
% of this training batch 
is then computed as
\begin{math}
\mathcal{L} = \frac{1}{2}\Sigma_{k=1}^b(\mathcal{L}^I_k + \mathcal{L}^T_k).
\end{math}
In this work, we explore two typical ways of measuring the similarity
between an image and a text. The learned representations of the image
and text are denoted as $\zi \in \mathbb{R}^{n_1 \times d}$ and
$\zt \in \mathbb{R}^{n_2 \times d}$, respectively. Here $n_1$ and
$n_2$ are the numbers of (non-padded) tokens in each image and text.

\textbf{Global Similarity.} In CLIP~\cite{radford2021learning} and
ALIGN~\cite{jia2021scaling}, the similarity is computed via dot
product of the global features of the entire image and text sequence.
Specifically, the global similarity between the image and text is
computed as
\begin{math}
   s^I_{i,j}  =   s^T_{i,j} = [\zi_i]_{\texttt{[CLS]}}^\top [\zt_j]_{\texttt{[SEP]}},
\end{math}
where $[\zi_i]_{\texttt{[CLS]}}$ denotes the feature vector of the
\texttt{[CLS]} token of the $i$-th image and
$[\zt_j]_{\texttt{[SEP]}}$ denotes the feature vector of the
\texttt{[SEP]} token of the $j$-th text. Since Swin Transformer has no
\texttt{[CLS]} token, we use the average pooling on the features of
all patch tokens to represent it.

\textbf{Token-wise Similarity.} In FILIP~\cite{yao2021filip}, the
similarity is computed based on a finer-grained interaction between
the image patches and textual tokens, which also brings good alignment and learns meaningful fine-grained features with
promising localization ability. For $i$-th image, each visual
token $[\zi_i]_k$ in it computes a similarity with all non-padded
textual tokens of the $j$-th text. Then the maximum one is used to
represent the token-wise similarity between this visual token and the
$j$-th text. Finally, we regard the average token-wise maximum
similarity of all non-padded tokens in this $i$-th image as the
cross-modal similarity
\begin{math}
  s_{i,j}^I   = \frac{1}{n_1}\sum_{k=1}^{n_1} [\zi_i]_k^\top [\zt_j]_{m_k^I},
\end{math}
where $m_k^I = \arg \max_{0\le r < n_2} [\zi_i]_k^\top [\zt_j]_r$.
The similarity of a text to an image can be computed in the same way, except that we exclude the \texttt{[CLS]}, \texttt{[SEP]}, and all padding tokens as in FILIP~\cite{yao2021filip}.

\textbf{Reduced-token Interaction.} Using the token-wise similarity
introduces a large amount of computation. The computation cost is about $2 \times n_1 \times n_2$ times more than
that of global similarity. The number of visual tokens $n_1$ is normally predefined while the number of textual tokens $n_2$ depends on the text input. To reduce the computation cost of token-wise similarity, an efficient way is to decrease the number of tokens involved in similarity calculation and we call this reduced-token
interaction.

In this paper, we propose a learnable token reduction layer on top of visual features output by the image encoder. The workflow of this layer is described in the right part of Figure~\ref{fig:model-arch}. Since the number of visual tokens is usually much larger than that of textual tokens, e.g., there are $16\times16+1=257$ visual tokens and 32 textual tokens for CLIP\textsubscript{ViT-L}, visual tokens are more necessary to be
decreased for efficiency. Denoting the visual tokens of an image sample as
$z^I \in \mathbb{R}^{n_1\times d}$, we aim to get a new
$Z^I = f(z^I) \in \mathbb{R}^{n^{\prime} \times d}$ in which $f$
denotes the function of token reduction and $n^{\prime}$ denotes the reduced token number. Finally, $z^I$ is replaced by $Z^I$ to calculate the token-wise similarity. In
general, given the output number of tokens $n^{\prime}$, the $k$-th
visual token $Z^I_k \in \mathbb{R}^d$ can be formulated by:
\begin{math}
  Z^I_k = \mathit{AvgPool}(\mathit{Conv}_k(z^I) \odot z^I), \quad k \in \{1, 2, \dots, n^{\prime}\}
\end{math}
where $\odot$ represents Hadamard product. Firstly,
$z^I_k \in \mathbb{R}^{n_1\times d}$ is reshaped to
$z^I_k \in \mathbb{R}^{H \times W \times d}$ in which $H$ and $W$
respectively represent the vertical and horizontal numbers of visual
tokens. Then, the $k$-th attention map is computed via
$\mathit{Conv_k}: \mathbb{R}^{H \times W \times d} \to \mathbb{R}^{H
  \times W \times 1}$ which is implemented using two
convolutional layers. We share the weight of $\mathit{Conv_k}$ across all $k$ tokens. Finally, a spatial global average pooling
$\mathit{AvgPool}: \mathbb{R}^{H \times W \times d} \to
\mathbb{R}^{d}$ is used to get the final $k$-th visual token.

\textbf{Locked-image Text tuning.} LiT-tuning~\cite{zhai2021lit} proposes that a locked pre-trained image encoder with an unlocked text encoder works well in contrastive learning. We extend this idea to cross-lingual data sources and try to align a locked image encoder pre-trained on
English data sources\added{, e.g., CLIP~\cite{radford2021learning} and FILIP~\cite{yao2021filip},} with a trainable Chinese text encoder. \added{These existing pre-trained image encoders usually have a projection linear layer. In our method, we drop this linear layer and add a new linear trainable random-initialized projection layer, whose output dimension
% in which case the output dimension of image encoder 
can be adjusted flexibly.}
Experiment results shown in Section~\ref{sec:ablation}
confirm its effectiveness.

% \vspace{-2mm}
% \subsection{Locked-image Text Tuning}
% \vspace{-2mm}
% Xiaohua Zhai et~al.~\cite{zhai2021lit} proposes that a locked pre-trained image encoder with unlocked text encoder works well in contrastive learning. Inspired by their work, 
% % we take advantage of pre-trained image encoders for contrastive learning. In particular, this setting aims to
% % teach a Chinese text encoder to read out suitable representations from
% % an existing image encoder pre-trained on English datasets.
% % techniques for learning image descriptors and vision-language
% % alignment~\cite{zhai2021lit}. And the image descriptors are
% % beforehand well pre-trained using relatively clean or (semi-) manually
% % labeled images.
% we extend this idea to
% % multilingual
% \jx{cross-lingual}
% data sources, and try to align a locked image encoder pre-trained on
% English data sources with a trainable Chinese text encoder.
% % Moreover,
% % the LiT-tuning method significantly speeds up the training process and
% % reduces memory requirements since no gradients need to be computed for
% % the visual encoder.
% Empirical results in Section~\ref{sec:ablation}
% confirm the effectiveness of our method.
%% Jiaxi & Xiaojun

%% chapter: Experiments
\vspace{-2mm}
\section{Wukong Chinese Benchmarks}\label{sec:expt}
\vspace{-2mm}
\subsection{Experimental Setup}
\vspace{-2mm}

% \textbf{Model Architectures and pretraining settings.}
Following the existing VLP models, \textit{e.g.,} CLIP~\cite{radford2021learning} and ALIGN~\cite{jia2021scaling}, we
employ a dual-encoder architecture as illustrated in
Figure~\ref{fig:model-arch}. 
We have three variations of pretraining Chinese models: global similarity (\textbf{CLIP}-style); token-wise similarity (\textbf{FILIP}-style) and token-wise similarity with token reduction layer (\textbf{Wukong}-style).
For different types of visual encoders, we have ViT-B, ViT-L\cite{dosovitskiy2020image}, and Swin-L\cite{liu2021swin}. 
\added{We use the token-wise similarity with our proposed reduced-token interaction for Wukong-style models.
For the dimension of the common multi-modal space, all the FILIP-style and Wukong-style models are set to 256 and CLIP-style models are set following the original CLIP checkpoints.} Models are trained using LiT-tuning~\cite{zhai2021lit}, since they achieve relatively better results as shown in Section~\ref{sec:ablation}. \added{In terms of pre-loaded visual encoders, CLIP and FILIP models with ViT-B/32 or ViT-L/14 are used. Swin-L pre-trained on ImageNet-22K with $224\times224$ image resolution is used for Swin Transformer based models, e.g., 
CLIP\textsubscript{Swin-L}.} \added{Detailed} training settings are in the appendix.

% Table~\ref{tab:model-arch}.

% In this way, we pre-train a few models with variants of visual
% encoders, on Wukong dataset. Among which, the largest in model size
% are Wukong\textsubscript{ViT} and Wukong\textsubscript{Swin},
% respectively using ViT-L/14 from FILIP~\cite{yao2021filip} and
% Swin-L from Swin Transformer~\cite{liu2021swin}. Details of model
% parameters and visual encoders are described in
% Table~\ref{tab:model-arch}.

% \xh{Put this in the Appendix}
% \textbf{Model Training Settings.} For better generalization and data-efficiency, we employ
% Autoaugment~\cite{cubuk2019autoaugment} for image data augmentation
% that aims to build more image-text pairs. All of our models are
% trained using Nvidia V100 GPUs and Ascend cards. Specifically,
% Wukong\textsubscript{ViT-B} is trained using 32 GPUs for 3 days, Wukong\textsubscript{ViT-L} is trained using 32 GPUs for 10 days and
% Wukong\textsubscript{Swin-L} is trained using 40 GPUs for 5
% days. We use LAMB optimizer~\cite{2019Large} and cosine
% learning rate schedule with a linear warmup~\cite{2016SGDR}. Weight
% decay regularization is applied to all parameters except for bias,
% layer normalization, token embedding, positional embedding and
% temperature in the contrastive loss. The detailed hyperparameters are
% shown in Appendix. In order to pick the optimal
% checkpoint out, ImageNet dataset~\cite{deng2009imagenet} with
% translated label names is used for zero-shot validation.

\vspace{-2mm}
\subsection{Zero-shot Image Classification}
\vspace{-2mm}

We evaluate our models for the zero-shot classification task
on 10 datasets whose class labels are translated from English. To make
the evaluation results more reliable, the translation process is done
with a machine translator and verified by human
experts. The Chinese annotations of these
datasets are released for future evaluation by the research community. Also, we evaluate BriVL~\cite{huo2021wenlan}, another multi-modal pre-training model for Chinese, on these datasets for zero-shot classification. The implementation code and pre-trained model weights of BriVL are both from its \href{https://github.com/chuhaojin/BriVL-BUA-applications/}{homepage}.

\textbf{Prompt Ensemble.} Text prompts are often used as a class label augmentation to achieve a
better performance in the zero-shot image classification
task~\cite{radford2021learning,yao2021filip}. For simplicity, instead of
designing prompts manually, we provide a set of 80 text prompts which are originally used on ImageNet by CLIP and manually translate them into Chinese. We also release these Chinese prompts for future fair comparison in our community. 

% It is possible that adding extra descriptions as a part of text prompts specific to each dataset may further improve the performance. 
% Since our paper aims to provide a benchmark dataset with pre-trained models, we stick to the
% general-purpose prompt templates and leave the design of better
% prompt templates for each task as a future work.

% Note that when using the global similarity, we follow CLIP to ensemble
% different prompt templates by using their mean textual representation,
% i.e., we sum different templates for the same class label to form a
% mean textual representation. While for the token-wise similarity, since there is no global representation, we
% simply ensemble prompt templates by their mean token-wise similarity
% to images.

\begin{table}
\vspace{-2mm}
  \setlength{\tabcolsep}{5pt}
  \centering
  \caption{Top-1 accuracy (\%) of the zero-shot image classification benchmark. All
    the models are trained using 100-million Wukong dataset except for BriVL which is pre-trained using its own dataset.
    % with token-wise similarity and reduced-token interaction method except
    % for those marked with special superscripts. The superscript ``F''
    % means models are trained using \textbf{F}ull-token token-wise
    % similarity and ``G'' means using \textbf{G}lobal
    % similarity.
    Results highlighted with \textbf{bold} mean the best within the same
    image encoder and those with \underline{underline} represent the best among all methods.}
  \label{tab:zero-shot} 
  \footnotesize  % use small font size in this table
  \begin{tabular}{p{2.8cm}|*{10}{c}|{c}}
    \diagbox[innerwidth=2.8cm]{Model}{Dataset (CN)} & \rotatebox[origin=c]{90}{CIFAR10} & \rotatebox[origin=c]{90}{CIFAR100} & \rotatebox[origin=c]{90}{Caltech101} & \rotatebox[origin=c]{90}{Caltech256} & \rotatebox[origin=c]{90}{DTD} & \rotatebox[origin=c]{90}{Sports} & \rotatebox[origin=c]{90}{Flowers} & \rotatebox[origin=c]{90}{SUN397} & \rotatebox[origin=c]{90}{EuroSAT} & \rotatebox[origin=c]{90}{ImageNet} & \rotatebox[origin=c]{90}{Average} \\
    \midrule
    BriVL~\cite{huo2021wenlan} & 72.3 & 35.9 & 72.0 & 58.0 & 18.8 & 83.6 & 18.4 & 28.4 & 25.5 & 24.3 & 43.72 \\
    \midrule[.1pt]
    CLIP\textsubscript{ViT-B}~\cite{radford2021learning} & \textbf{89.4} & 62.5 & \textbf{89.2} & \textbf{82.7} & 36.2 & 93.1 & 52.6 & 55.8 & 25.7 & 47.7 & 63.49 \\
    FILIP\textsubscript{ViT-B}~\cite{yao2021filip} & 87.0 & 53.3 & 83.1 & 71.0 & 28.9 & 91.2 & 48.8 & 50.0 & 29.5 & 38.1 & 58.09 \\
    Wukong\textsubscript{ViT-B} & 87.1 & \textbf{62.6} & 89.1 & 82.3 & \textbf{37.3} & \textbf{95.6} & \textbf{64.8} & \textbf{56.0} & \textbf{32.6} & \textbf{49.1} & \textbf{65.65} \\
    \midrule[.2pt]
    CLIP\textsubscript{ViT-L}~\cite{radford2021learning} & 94.1 & 71.3 & 91.9 & 89.0 & 45.4 & 98.7 & \textbf{72.3} & \underline{\textbf{62.6}} & 42.8 & \textbf{57.9} & 72.60 \\
    FILIP\textsubscript{ViT-L}~\cite{yao2021filip} & 90.6 & 66.3 & 89.9 & 86.2 & \underline{\textbf{46.4}} & 97.8 & 69.4 & 60.2 & 25.5 & 54.0 & 68.63 \\
    Wukong\textsubscript{ViT-L} & \textbf{95.4} & \textbf{77.1} & \underline{\textbf{92.4}} & \underline{\textbf{89.2}} & 40.9 & \underline{\textbf{99.1}} & 68.9 & 62.0 & \underline{\textbf{50.3}} & 55.0 & \underline{\textbf{73.03}} \\
    \midrule[.2pt]
    CLIP\textsubscript{Swin-L}~\cite{radford2021learning} & 94.8 & 75.8 & 90.7 & 88.3 & \textbf{40.0} & 97.5 & 71.0 & \textbf{57.3} & \textbf{22.3} & 58.0 & 69.57 \\
    FILIP\textsubscript{Swin-L}~\cite{yao2021filip} & \underline{\textbf{95.5}} & \underline{\textbf{77.2}} & \textbf{91.6} & \textbf{88.4} & 39.8 & \underline{\textbf{99.1}} & 75.1 & 56.5 & 21.0 & \underline{\textbf{58.5}} & \textbf{70.27} \\
    Wukong\textsubscript{Swin-L} & 95.3 & 76.8 & 89.8 & 87.1 & 33.7 & 97.8 & \underline{\textbf{76.9}} & 56.3 & 19.3 & 58.2 & 69.12 \\
    \bottomrule
  \end{tabular}
  \vspace{-6mm}
\end{table}

\textbf{Performance.} The evaluation of zero-shot image classification on different datasets
is illustrated in Table~\ref{tab:zero-shot}. In addition to our proposed models, i.e., Wukong\textsubscript{ViT-B}, Wukong\textsubscript{ViT-L}, and Wukong\textsubscript{Swin-L}, we also evaluate other model architectures, i.e., CLIP and FILIP, with different image encoders as comparisons. These models are all pre-trained using our Wukong dataset except for BriVL which uses its own dataset. In comparison with models pre-trained using Wukong dataset, BriVL shows a significantly poor performance. This can be considered as the proof that Wukong dataset is effective for multi-modal pre-training. Besides, using the same ViT image encoder, either ViT-B or ViT-L, Wukong models perform quite well. In particular, Wukong\textsubscript{ViT-L} achieves the highest average accuracy of 73.03\% among all models. This indicates the superiority of our model architecture. However, our model trained with SwinT as the image encoder performs worse compared to others. The reason might be that patch merging in SwinT has already served a similar purpose in selecting and merging the important visual patch tokens. Therefore, our reduced-token interaction brings a negative impact. 
In summary, the zero-shot classification performances on various tasks show the effectiveness of our dataset and Wukong models. 
\vspace{-2mm}
\subsection{Image-Text Retrieval}
\vspace{-2mm}
In this section, we evaluate our models on two sub-tasks, including
image-to-text retrieval and text-to-image retrieval. In the
image-to-text retrieval, the model retrieves a target text from a set
of candidates given an image as query, or vice versa for the
text-to-image retrieval. We benchmark our models on 6 different
datasets, including Flickr8K-CN~\cite{flickr8k-cn},
Flickr30K-CN~\cite{flickr30k-cn}, COCO-CN~\cite{coco-cn},
AIC-ICC~\cite{aicicc}, MUGE\footnote{\url{https://tianchi.aliyun.com/muge}}
and Wukong-Test.
% \jx{As a comparison, we also evaluate BriVL~\cite{huo2021wenlan} on these datasets for image-text retrieval.}

% For each dataset, our pre-trained models are evaluated in both
% zero-shot and fine-tuned settings \xj{(need to mention zero-shot in Appendix)}.
Following common practices, we
report Recall@K (recall of top K candidates) with $K = 1,5,10$ for
both image-to-text and text-to-image retrieval on all datasets except for
MUGE, which only has the text-to-image retrieval setting. The average
Recall@K, i.e., Mean Recall (MR), is used for the final comparison.
We report results on the test sets, except for MUGE and AIC-ICC where
test sets are not released. For MUGE, we report results on the
validation set, and for AIC-ICC, following the setting of
WenLan~2.0~\cite{fei2021wenlan}, we take the first 10K images along
with their corresponding 50K pieces of texts from the validation set for testing. 
%We compare our models against several baseline methods.

% \subsection{Zero-shot Image-text Retrieval}

\begin{table}[!t]
  \vspace{-4mm}
  \setlength{\tabcolsep}{1pt}
  \renewcommand{\arraystretch}{0.7}
  \centering
  \caption{Benchmarks of zero-shot image-text retrieval. The top-3 performance values are highlighted with \tfbestfirst{bold}, \tfbestsecond{underline} and \tfbestthird{italic} respectively.}
  \label{tab:retrieval-zero-shot}
  \footnotesize  % use small font size in this table
  \begin{tabularx}{\textwidth}{cl *3{>{\centering}X} *3{>{\centering}X} c}
    \toprule
    \multirow{2}{*}{Dataset} & \multicolumn{1}{c}{\multirow{2}{*}{Method}} & \multicolumn{3}{c}{Image-to-Text Retrieval} & \multicolumn{3}{c}{Text-to-Image Retrieval} & \multirow{2}{*}{MR} \\
    \cmidrule(lr){3-5}\cmidrule(lr){6-8}
    & \multicolumn{1}{c}{} & R@1 & R@5 & R@10 & R@1 & R@5 & R@10 & \\
    \midrule \multirow{10}{*}{Flickr8K-CN}
    & BriVL~\cite{huo2021wenlan} & 13.4 & 31.2 & 40.7 & 8.0 & 20.7 & 29.5 & 23.9 \\
    & CLIP\textsubscript{ViT-B} & 59.5 & 86.2 & 93.4 & 44.2 & 71.2 & 82.0 & 72.7 \\
    & CLIP\textsubscript{ViT-L}~\cite{radford2021learning} & 65.4 & 89.2 & 95.4 & 50.5 & 77.0 & 85.7 & \tfbestsecond{77.2} \\
    & CLIP\textsubscript{Swin-L} & 56.0 & 83.2 & 92.4 & 38.6 & 67.0 & 78.2 & 69.2 \\
    & FILIP\textsubscript{ViT-B} & 37.2 & 65.9 & 75.2 & 24.0 & 50.0 & 62.4 & 52.5 \\
    & FILIP\textsubscript{ViT-L}~\cite{yao2021filip} & 70.0 & 91.6 & 96.6 & 53.5 & 79.3 & 87.9 & \tfbestfirst{79.8} \\
    & FILIP\textsubscript{Swin-L} & 52.4 & 78.0 & 87.2 & 41.2 & 68.5 & 79.1 & 67.7 \\
    & Wukong\textsubscript{ViT-B} & 55.4 & 82.3 & 90.0 & 43.2 & 71.3 & 81.3 & 70.6 \\
    & Wukong\textsubscript{ViT-L} & 61.4 & 86.2 & 93.6 & 46.0 & 74.5 & 84.5 & \tfbestthird{74.4} \\
    & Wukong\textsubscript{Swin-L} & 47.2 & 78.8 & 87.6 & 36.6 & 64.8 & 76.2 & 65.2 \\
    \midrule[.2pt]
    \multirow{10}{*}{Flickr30K-CN} & BriVL~\cite{huo2021wenlan} & 17.7 & 42.3 & 54.3 & 10.3 & 27.5 & 37.9 & 31.7 \\
    & CLIP\textsubscript{ViT-B} & 72.2 & 92.0 & 96.4 & 47.2 & 74.1 & 82.9 & 77.5 \\
    & CLIP\textsubscript{ViT-L}~\cite{radford2021learning} & 75.0 & 94.5 & 97.7 & 51.8 & 78.6 & 85.9 & \tfbestthird{80.6} \\
    & CLIP\textsubscript{Swin-L} & 64.3 & 89.3 & 94.3 & 41.2 & 69.7 & 80.2 & 73.2 \\
    & FILIP\textsubscript{ViT-B} & 44.2 & 73.7 & 83.3 & 28.7 & 55.9 & 67.1 & 58.8 \\
    & FILIP\textsubscript{ViT-L}~\cite{yao2021filip} & 78.9 & 96.2 & 98.1 & 55.7 & 81.2 & 87.9 & \tfbestfirst{83.0} \\
    & FILIP\textsubscript{Swin-L} & 65.8 & 89.2 & 95.0& 44.6 & 72.2& 81.2& 74.7 \\
    & Wukong\textsubscript{ViT-B} & 66.2 & 88.7 & 94.3 & 45.7 & 73.8 & 82.2 & 75.1 \\
    & Wukong\textsubscript{ViT-L} & 76.1 & 94.8 & 97.5 & 51.7 & 78.9 & 86.3 & \tfbestsecond{80.9} \\
    & Wukong\textsubscript{Swin-L} & 58.7 & 86.7 & 92.7 & 40.9 & 68.0 & 78.4 & 70.9 \\
    \midrule[.2pt]
    \multirow{10}{*}{COCO-CN} & BriVL~\cite{huo2021wenlan} & 17.1 & 41.7 & 57.5 & 14.8 & 39.0 & 54.2 & 37.4 \\
    & CLIP\textsubscript{ViT-B} & 52.8 & 79.6 & 88.9 & 48.7 & 79.4 & 88.5 & \tfbestthird{73.0} \\
    & CLIP\textsubscript{ViT-L}~\cite{radford2021learning} & 51.0 & 80.0 & 89.7 & 48.7 & 76.8 & 86.4 & 72.1 \\
    & CLIP\textsubscript{Swin-L} & 50.5 & 79.2 & 88.2 & 46.7 & 78.1 & 87.7 & 71.7 \\
    & FILIP\textsubscript{ViT-B} & 37.8 & 66.4 & 77.9 & 37.5 & 68.1 & 83.0 & 61.8 \\
    & FILIP\textsubscript{ViT-L}~\cite{yao2021filip} & 56.9 & 82.4 & 90.9 & 52.7 & 79.9 & 88.6 & \tfbestfirst{75.2} \\
    & FILIP\textsubscript{Swin-L} & 48.6 & 77.3 & 88.3 & 50.5 & 79.2 & 88.6 & 72.1 \\
    & Wukong\textsubscript{ViT-B} & 48.3 & 77.8 & 88.8 & 49.2 & 79.4 & 87.9 & 71.9 \\
    & Wukong\textsubscript{ViT-L} & 55.2 & 81.0 & 90.6 & 53.4 & 80.2 & 90.1 & \tfbestsecond{75.1} \\
    & Wukong\textsubscript{Swin-L} & 47.3 & 78.0 & 88.3 & 46.4 & 77.0 & 87.6 & 70.8 \\
    \midrule[.2pt]
    \multirow{10}{*}{MUGE}    & BriVL~\cite{huo2021wenlan} & - & - & - & 12.7 & 30.9 & 41.8 & 28.5 \\
    & CLIP\textsubscript{ViT-B} & - & - & - & 37.3 & 64.2 & 73.9 & \tfbestthird{58.5} \\
    & CLIP\textsubscript{ViT-L}~\cite{radford2021learning} & - & - & - & 43.3 & 69.2 & 78.4 & \tfbestfirst{63.6} \\
    & CLIP\textsubscript{Swin-L} & - & - & - & 35.2 & 62.2 & 73.2 & 56.9 \\
    & FILIP\textsubscript{ViT-B} & - & - & - & 22.4 & 46.6 & 58.5 & 42.5 \\
    & FILIP\textsubscript{ViT-L}~\cite{yao2021filip} & - & - & - & 37.6 & 63.4 & 73.6 & 58.2 \\
    & FILIP\textsubscript{Swin-L} & - & - & - & 36.2 & 61.1 & 71.5 & 56.3 \\
    & Wukong\textsubscript{ViT-B} & - & - & - & 33.4 & 59.3 & 69.7 & 54.1 \\
    & Wukong\textsubscript{ViT-L} & - & - & - & 42.7 & 69.0 & 78.0 & \tfbestsecond{63.2} \\
    & Wukong\textsubscript{Swin-L} & - & - & - & 34.5 & 60.6 & 71.2 & 55.5 \\
    \bottomrule
  \end{tabularx}
\vspace{-3mm}
\end{table}

\added{Table~\ref{tab:retrieval-zero-shot} shows the benchmarks of zero-shot image-text retrieval using different models on multiple datasets. In general, models trained on Wukong dataset achieve a significantly better performance than BriVL~\cite{huo2021wenlan}, which demonstrates the effectiveness of our dataset.
Besides, Wukong\textsubscript{ViT-L} shows a competitive performance in comparison to other models. Therefore, we believe Wukong dataset can serve as a pre-training benchmark dataset with a wide coverage of concepts.}

\begin{table}[!t]
\vspace{-4mm}
  \setlength{\tabcolsep}{1pt}
  \renewcommand{\arraystretch}{0.7}
  \centering
\caption{Benchmarks of fine-tuned image-text retrieval on different
    datasets. The top-3 performance values are highlighted with \tfbestfirst{bold}, \tfbestsecond{underline} and \tfbestthird{italic} respectively.}
    \label{tab:retrieval-finetuned}
  \footnotesize  % use small font size in this table
    \begin{tabularx}{\textwidth}{>{\centering\bfseries}m{2.4cm} l *3{>{\centering}X} *3{>{\centering}X} c}
      \toprule
      \multirow{2}{*}{Dataset} & \multicolumn{1}{c}{\multirow{2}{*}{Method}} & \multicolumn{3}{c}{Image-to-Text Retrieval} & \multicolumn{3}{c}{Text-to-Image Retrieval} & \multirow{2}{*}{MR} \\
      \cmidrule(lr){3-5}\cmidrule(lr){6-8}
      & & R@1 & R@5 & R@10 & R@1 & R@5 & R@10 & \\
      \midrule[.2pt]
      \multirow{10}{*}{Flickr8K-CN}
      & CLIP\textsubscript{ViT-B} & 77.7 & 94.7 & 98.1 & 61.2 & 86.8 & 93.2 & 85.3 \\
      & CLIP\textsubscript{ViT-L}~\cite{radford2021learning} & 81.4 & 96.9 & 99.0 & 67.4 & 91.0 & 95.7 & \tfbestsecond{88.6} \\
      & CLIP\textsubscript{Swin-L} & 77.3 & 94.9 & 98.2 & 59.3 & 86.0 & 92.9 & 84.8 \\
      & FILIP\textsubscript{ViT-B} & 52.6 & 81.5 & 90.2 & 46.4 & 77.0 & 86.8 & 72.4 \\
      & FILIP\textsubscript{ViT-L}~\cite{yao2021filip} & 80.8 & 94.8 & 98.3 & 68.5 & 90.5 & 95.2 & \tfbestthird{88.0} \\
      & FILIP\textsubscript{Swin-L} & 77.6 & 94.4 & 97.7 & 61.5 & 86.5 & 93.0 & 85.1 \\
      & Wukong\textsubscript{ViT-B} & 71.7 & 91.5 & 96.6 & 58.4 & 85.4 & 92.0 & 82.6 \\
    %   & Wukong\textsubscript{ViT-L} (from-scratch) & 2.2 & 8.6 & 13.2 & 2.0 & 6.6 & 11.0 & 7.3 \\
      & Wukong\textsubscript{ViT-L} & 83.3 & 97.3 & 99.5 & 70.1 & 91.9 & 96.4 & \tfbestfirst{89.7} \\
      & Wukong\textsubscript{Swin-L} & 74.9 & 93.6 & 97.8 & 57.9 & 85.1 & 92.6 & 83.6 \\
      \midrule[.2pt]
      \multirow{10}{*}{Flickr30K-CN}
      & CLIP\textsubscript{ViT-B} & 87.1 & 97.7 & 98.8 & 69.0 & 90.3 & 95.0 & 89.7 \\
      & CLIP\textsubscript{ViT-L}~\cite{radford2021learning} & 91.6 & 99.1 & 99.7 & 77.3 & 94.4 & 97.2 & \tfbestsecond{93.2} \\
      & CLIP\textsubscript{Swin-L} & 85.8 & 97.1 & 99.0 & 67.4 & 90.3 & 94.9 & 89.1 \\
      & FILIP\textsubscript{ViT-B} & 72.1 & 91.3 & 95.8 & 57.5 & 84.3 & 90.6 & 81.9 \\
      & FILIP\textsubscript{ViT-L}~\cite{yao2021filip} & 90.6 & 98.8 & 99.6 & 76.9 & 94.9 & 97.4 & \tfbestthird{93.0} \\
      & FILIP\textsubscript{Swin-L} & 86.0 & 97.5 & 99.1 & 70.9 & 91.3 & 95.3 & 90.0 \\
      & Wukong\textsubscript{ViT-B} & 83.9 & 97.6 & 99.0 & 67.6 & 89.6 & 94.2 & 88.7 \\
    %   & Wukong\textsubscript{ViT-L} (from-scratch) & 5.8 & 17.3 & 25.4 & 3.8 & 13.2 & 20.6 & 14.4 \\
      & Wukong\textsubscript{ViT-L} & 92.7 & 99.1 & 99.6 & 77.4 & 94.5 & 97.0 & \tfbestfirst{93.4} \\
      & Wukong\textsubscript{Swin-L} & 86.2 & 98.1 & 99.4 & 67.4 & 89.9 & 94.5 & 89.3 \\
      \midrule[.2pt]
      \multirow{17}{*}{COCO-CN} & EmbN~\cite{wang2018learning} & - & - & - & - & - & - & 73.2 \\
      & PARALLEL-EmbN~\cite{gella2017image} & - & - & - & - & - & - & 76.0 \\
      & S-LIWE~\cite{wehrmann2019language} & - & - & - & - & - & - & 73.6 \\
      & M\textsuperscript{3}P~\cite{ni2021m3p} & - & - & - & - & - & - & 86.2 \\
      & UNITER~\cite{chen2020uniter} & - & - & - & - & - & - & 87.3 \\
      & LightningDOT~\cite{sun2021lightningdot} & - & - & - & - & - & - & \tfbestthird{88.4} \\
      & UC\textsuperscript{2}~\cite{zhou2021uc2} & - & - & - & - & - & - & \tfbestfirst{89.8} \\
      & CLIP\textsubscript{ViT-B} & 68.7 & 93.6 & 97.5 & 68.9 & 93.3 & 97.3 & 86.6 \\
      & CLIP\textsubscript{ViT-L}~\cite{radford2021learning} & 68.3 & 93.0 & 97.3 & 70.1 & 92.2 & 96.4 & 86.2 \\
      & CLIP\textsubscript{Swin-L} & 68.0 & 92.8 & 97.3 & 66.7 & 91.5 & 96.3 & 85.4 \\
      & FILIP\textsubscript{ViT-B} & 52.7 & 81.3 & 88.3 & 56.2 & 86.8 & 94.3 & 76.6 \\
      & FILIP\textsubscript{ViT-L}~\cite{yao2021filip} & 69.1 & 91.3 & 96.9 & 72.2 & 92.4 & 97.2 & 86.5 \\
      & FILIP\textsubscript{Swin-L} & 68.3 & 93.9 & 97.1 & 69.9 & 93.3 & 97.6 & 86.7 \\
      & Wukong\textsubscript{ViT-B} & 65.8 & 90.3 & 96.6 & 67.0 & 91.4 & 96.7 & 84.6 \\
    %   & Wukong\textsubscript{ViT-L} (from-scratch) & 1.8 & 10.6 & 17.0 & 2.8 & 8.4 & 15.2 & 9.3 \\
      & Wukong\textsubscript{ViT-L} & 73.3 & 94.0 & 98.0 & 74.0 & 94.4 & 98.1 & \tfbestsecond{88.6} \\
      & Wukong\textsubscript{Swin-L} & 67.4 & 92.4 & 97.5 & 66.0 & 92.6 & 97.1 & 85.5 \\
      \midrule[.2pt]
      \multirow{11}{*}{AIC-ICC} & WenLan~2.0~\cite{fei2021wenlan} & 45.6 & 68.0 & 76.3 & 34.1 & 58.9 & 69.1 & 58.7 \\
      & CLIP\textsubscript{ViT-B} & 50.5 & 73.0 & 80.2 & 38.1 & 63.7 & 73.3 & 63.1 \\
      & CLIP\textsubscript{ViT-L}~\cite{radford2021learning} & 59.1 & 79.5 & 85.2 & 46.2 & 70.7 & 78.6 & \tfbestsecond{69.9} \\
      & CLIP\textsubscript{Swin-L} & 50.5 & 73.5 & 81.2 & 37.3 & 62.8 & 72.7 & 63.0 \\
      & FILIP\textsubscript{ViT-B} & 42.5 & 67.2 & 76.0 & 32.9 & 58.4 & 68.8 & 57.6 \\
      & FILIP\textsubscript{ViT-L}~\cite{yao2021filip} & 54.1 & 75.8 & 82.8 & 44.9 & 69.0 & 77.5 & \tfbestthird{67.4} \\
      & FILIP\textsubscript{Swin-L} & 53.1 & 74.8 & 82.0 & 41.1 & 65.7 & 74.7 & 65.2 \\
      & Wukong\textsubscript{ViT-B} & 47.5 & 70.6 & 78.6 & 36.7 & 36.7 & 71.7 & 57.0 \\
      & Wukong\textsubscript{ViT-L} & 61.6 & 80.5 & 86.1 & 48.6 & 72.5 & 80.2 & \tfbestfirst{71.6} \\
      & Wukong\textsubscript{Swin-L} & 50.9 & 73.6 & 81.5 & 38.6 & 64.1 & 73.6 & 63.7 \\
      \midrule[.2pt]
      \multirow{10}{*}{MUGE}
      & CLIP\textsubscript{ViT-B} & - & - & - & 43.5 & 71.7 & 80.6 & 65.3 \\
      & CLIP\textsubscript{ViT-L}~\cite{radford2021learning} & - & - & - & 50.1 & 76.9 & 84.9 & \tfbestsecond{70.6} \\
      & CLIP\textsubscript{Swin-L} & - & - & - & 45.3 & 72.1 & 81.1 & \tfbestthird{66.2} \\
      & FILIP\textsubscript{ViT-B} & - & - & - & 30.6 & 58.2 & 70.2 & 53.0 \\
      & FILIP\textsubscript{ViT-L}~\cite{yao2021filip} & - & - & - & 43.5 & 71.5 & 80.9 & 65.3 \\
      & FILIP\textsubscript{Swin-L} & - & - & - & 44.0 & 71.4 & 81.2 & 65.5 \\
      & Wukong\textsubscript{ViT-B} & - & - & - & 39.2 & 66.9 & 77.4 & 61.2 \\
      & Wukong\textsubscript{ViT-L} & - & - & - & 52.7 & 77.9 & 85.6 & \tfbestfirst{72.1} \\
      & Wukong\textsubscript{Swin-L} & - & - & - & 43.8 & 71.9 & 81.7 & 65.8 \\
      \midrule[.2pt]
    \multirow{10}{*}{Wukong-Test} 
    % & BriVL~\cite{huo2021wenlan} & 18.3 & 40.5 & 51.4 & 17.9 & 38.9 & 49.4 & 36.1 \\
    & CLIP\textsubscript{ViT-B} & 58.3 & 88.2 & 94.1 & 53.1 & 85.4 & 92.6 & \tfbestthird{78.6} \\
    & CLIP\textsubscript{ViT-L}~\cite{radford2021learning} & 72.8 & 98.2 & 99.8 & 68.9 & 98.0 & 99.8 & \tfbestfirst{89.6} \\
    & CLIP\textsubscript{Swin-L} & 56.0 & 86.1 & 92.5 & 51.0 & 83.4 & 90.9 & 76.7 \\
    & FILIP\textsubscript{ViT-B} & 30.3 & 57.6 & 66.9 & 20.2 & 47.5 & 60.3 & 47.1 \\
    & FILIP\textsubscript{ViT-L}~\cite{yao2021filip} & 53.0 & 85.3 & 92.7 & 50.4 & 84.1 & 92.0 & 76.3 \\
    & FILIP\textsubscript{Swin-L} & 51.0 & 81.6 & 88.9 & 45.2 & 77.9 & 87.0 & 71.9 \\
    & Wukong\textsubscript{ViT-B} & 50.5 & 82.7 & 90.5 & 47.1 & 80.1 & 88.9 & 73.3 \\
    & Wukong\textsubscript{ViT-L} & 68.0 & 94.4 & 98.0 & 63.8 & 93.0 & 97.3 & \tfbestsecond{85.8} \\
    & Wukong\textsubscript{Swin-L} & 53.1 & 85.4 & 92.2 & 47.8 & 81.6 & 89.7 & 75.0 \\
      \bottomrule
    \end{tabularx}
\vspace{-3mm}
\end{table}

Table~\ref{tab:retrieval-finetuned} shows the results of image-text retrieval task. 
Generally, Wukong\textsubscript{ViT-L} achieves the best results among different model variants and datasets.
% \jx{Generally, Wukong\textsubscript{ViT-L} performs well in either zero-shot or fine-tuned image-text retrieval task. Relatively, our model gets more competitive results in fine-tuned settings. One explanation might be that token reduction layer extracts fewer tokens for cross-modality interaction which are not expressive enough without fine-tuning.}
Compared with baseline methods, on AIC-ICC, Wukong significantly outperforms WenLan~2.0
by around 12.9\%, which was pre-trained on a larger dataset
consisting of 650 million image-text pairs. For the COCO-CN dataset,
our Wukong models also achieve comparable performance to
state-of-the-art methods. For Wukong-Test, CLIP\textsubscript{ViT-L} achieves the best result (89.6\%) so far. It shows that models with global similarity is particularly effective when massively trained on in-domain Wukong train set. However, it lacks a bit of generalization when finetuned on other out-of-domain datasets such as AIC-ICC and MUGE.
% In addition, Wukong models also perform well in benchmarks of zero-shot retrieval, and detailed results can be found in the appendix.
% (Benchmarks of the zero-shot retrieval can be found in the appendix.)
Overall, experimental results demonstrate the capabilities of our pre-trained models.
% \added{Moreover, to justify the effect of Wukong dataset for multi-modal pre-training, we also evaluate Wukong\textsubscript{ViT-L} without loading pre-trained model weights, i.e., training from scratch. From the results in Table~\ref{tab:retrieval-finetuned}, We can find that models with pre-trained model weights loaded significantly outperform those without pre-training. We can see from such results that pre-training on Wukong provides a good understanding of visual and textual concepts.}

\vspace{-2mm}
\subsection{Ablations and Findings}
\label{sec:ablation}
\vspace{-2mm}

\begin{figure}
\vspace{-2mm}
    \centering
    \begin{subfigure}[b]{0.40\textwidth}
        \centering
        \includegraphics[width=\textwidth]{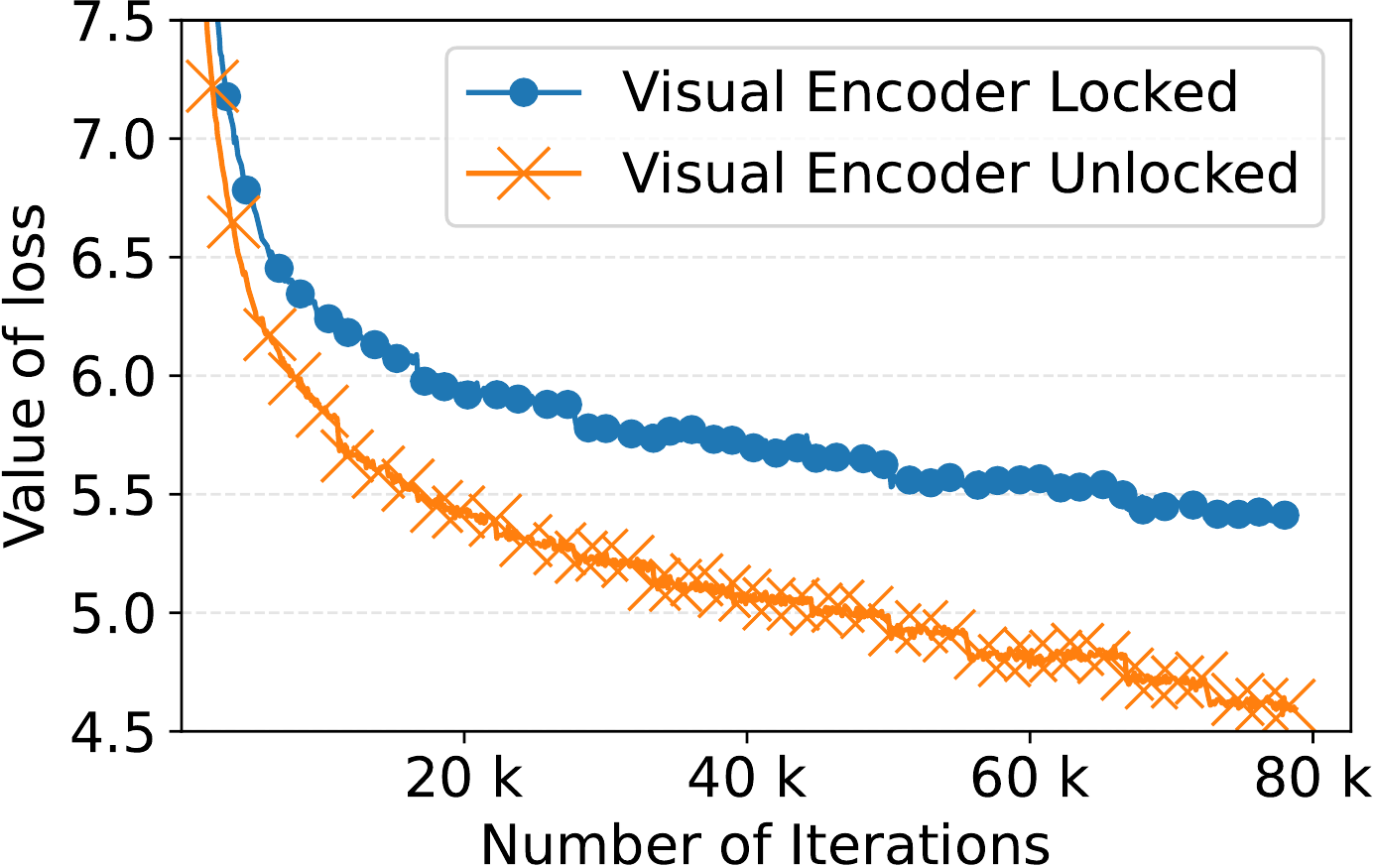}
        \caption{Training loss.}
        % \caption{The loss changes in training.}
    \end{subfigure}
    \hfill
    \begin{subfigure}[b]{0.40\textwidth}
        \centering
        \includegraphics[width=\textwidth]{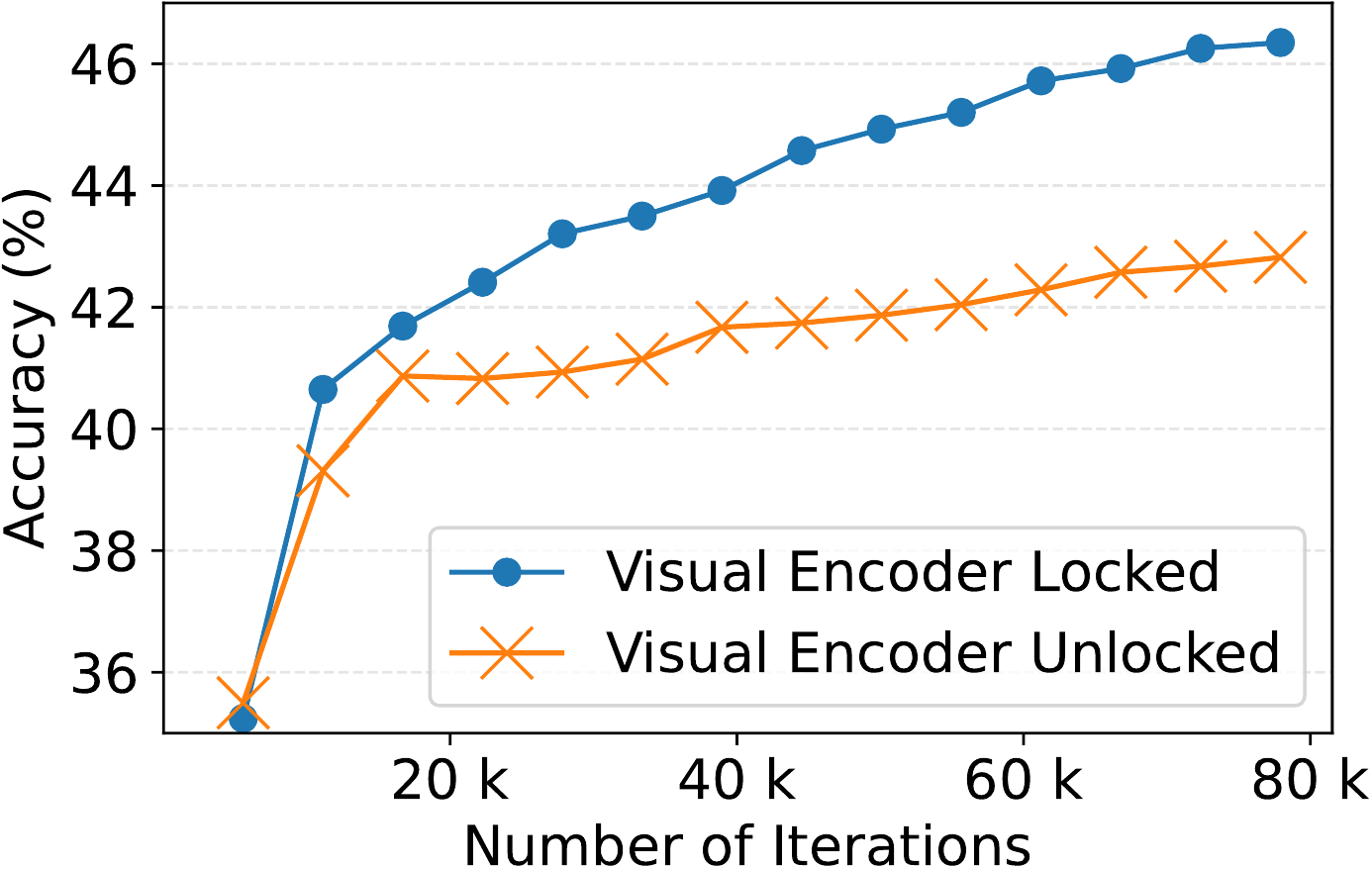}
        % \caption{The accuracy changes in validation.}
        \caption{Validation accuracy.}
    \end{subfigure}
    \vspace{-1mm}
    \caption{\label{fig:ablation_lit} In comparison with the model trained with an unlocked image encoder, though the loss decreases slower when the image encoder is locked, the accuracy of evaluation remains a higher level.
    % in every iteration
    }
\vspace{-4mm}
\end{figure}

\textbf{Locked-image Text Tuning.} To evaluate the effectiveness of LiT-tuning, we take Wukong\textsubscript{ViT-B} as an example model for a detailed investigation. We train two models using the same experimental settings as mentioned above, apart from that one model is trained with a locked image encoder but the other is not locked.
%The dataset for training is Wukong and that for evaluation is ImageNet with Chinese label names. 
% Figure~\ref{fig:ablation_lit} shows the changes of metrics of these two models. During training, 
As shown in Figure~\ref{fig:ablation_lit}, the model using LiT-tuning method
%with image encoder locked 
shows a slower trend of loss decrease during training. We believe the unlocked image encoder contributes to reduce the training loss and find the local optima efficiently. However, the validation accuracy of LiT-tuning model remains higher  than the other in almost every iteration, which demonstrates a better generalization.
%As a result, training with a locked image encoder has better generalization.

\textbf{Visualization.} In addition, we present the visualization of word-patch alignment in the appendix, which evidences the effectiveness of cross-modal token-wise similarity even in the LiT-tuning setting. We apply the same visualization method from FILIP~\cite{yao2021filip}, to align textual tokens and image patch tokens from FILIP\textsubscript{ViT-L} and FILIP\textsubscript{Swin-L}.
% As shown in the Appendix, Figure~\ref{fig:vis}, 
We find that both models can predict image patches of the target object, and more details are shown in the appendix.
% Though the visual encoder is locked in the pre-training phase, the learnable linear projection layer on top of it is still able to align patches and words in a fine-grained manner.
% We also find that this token-wise similarity in the loss function
% works across various patch-based visual encoders, not only the
% traditional ViT architecture but also SwinT that uses the
% sophisticated shifted windowing attention and patch merging.
Given this promising capability of aligning words and patches, our released models offer a potential solution for image object
localization. 
%Details of the visualization are shown in the appendix.

\textbf{Tokenization for Chinese.} We investigate the influence of the word segmentation technique on Chinese VLP models. Comparing the common character-grained tokenization, word-grained tokenization with a larger vocabulary (\num{65328}) is also adopted. Results show that the model using character-grained tokenization achieves better performance. The detailed comparison is shown in the appendix. Since a Chinese word often contains more than one character, the character-grained tokens are more fine-grained than word-grained. One example is that the word ``\texttt{蜂鸟}''(hummingbird) consists of two characters: ``\texttt{蜂}''(bee) and ``\texttt{鸟}'' (bird). Therefore, we believe it is more effective for our models to learn deep semantic token-wise similarity between an image patch and its paired fine-grained textual tokens, in such a contrastive learning manner.

% \textbf{Tokenization for Chinese.} \xj{To investigate the effect of Chinese word segmentation on Wukong model performance, we use the common python module \textit{jieba} to perform Chinese word segmentation to split Chinese text into words, rather than the presented character-tokenized technique. This new Chinese word-tokenized vocabulary size is 65328 that is nearly three times as much as the character-tokenized one. Results from Table~\ref{tab:char-word-token} show that Wukong\textsubscript{ViT-B} achieve better performance than Wukong\textsubscript{ViT-B-Word}. Since a Chinese word often contains more than one character, the character-tokenized textual tokens are more fine-grained than word-tokenized. One example is that the word ``\texttt{蜂鸟}''(hummingbird) consists of two characters ``\texttt{蜂}''(bee) and ``\texttt{鸟}'' (bird). Therefore, we believe it is more effective for Wukong\textsubscript{ViT-B} to learn deep semantic token-wise similarity between an image patch and its paired fine-grained textual tokens.}

% \paragraph{Prompts for Chinese} TODO (maybe put in appendix)

%% Jiaxi & Xiaojun

%% chapter: Conclusion
\vspace{-2mm}
\section{Conclusion}
\vspace{-2mm}

In this work, we build a large-scale Chinese vision-language dataset
called Wukong. To the best of our knowledge, it is the first
hundred-million level dataset designed for the Chinese language and it
paves the way for future research on Chinese cross-modal
pre-training. Meanwhile, using this dataset, we propose three Chinese
VLP models, i.e., Wukong\textsubscript{ViT-B},
Wukong\textsubscript{ViT-L}, and
Wukong\textsubscript{Swin-L}. Our pre-trained
Wukong\textsubscript{ViT-L} achieves state-of-the-art performance on
Chinese benchmarks such as zero-shot image classification and
image-text retrieval tasks.  In the future, we plan to explore more
solutions to train multilingual cross-modal models with the Wukong
dataset. \added{Meanwhile, more downstream tasks, in addition to 
image classification and retrieval, are worth sufficient evaluation.
Also, Wukong-based applications such as image search engines and visual question answering will be further explored in future work.}

% \begin{ack}
%   [TODO] Acknowledgments here.
% \end{ack}

%% chapter: Reference
\newpage

{\small \bibliography{ms}}

%% chapter: Checklist for submission
% \input{contents/checklist}

%% chapter: Appendix
\vspace{4mm}
\textbf{\Large Appendix}
\appendix
\section{Examples in Wukong Dataset}

\begin{figure}[!h]
  \centering \includegraphics[width=.95\linewidth]{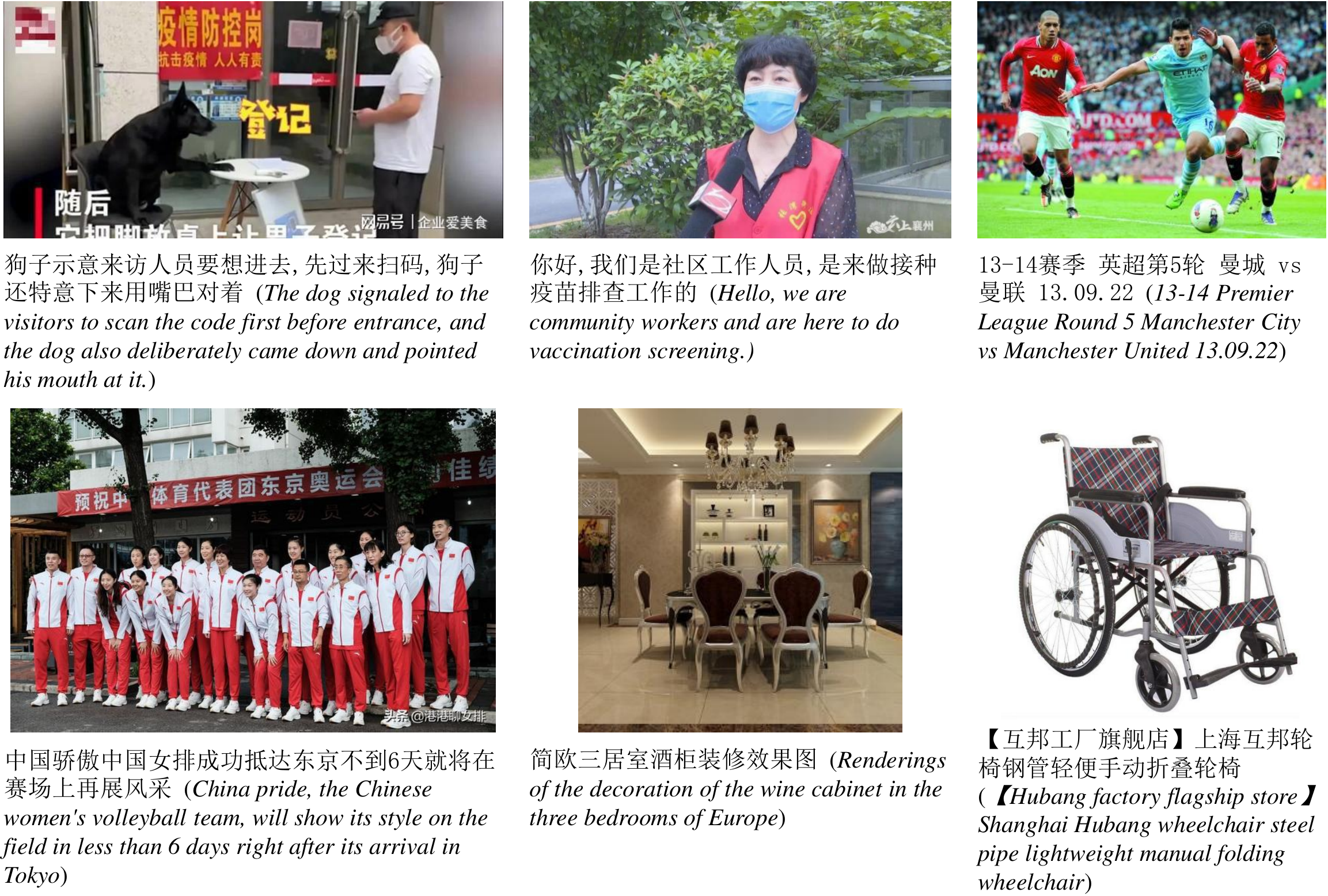}
  \caption{Examples of image-text pairs in Wukong dataset. A 
  diverse range of concepts are included.}
  \label{fig:dataset_samples}
\end{figure}

\begin{figure}[!h]
  \centering
  \includegraphics[width=.8\linewidth]{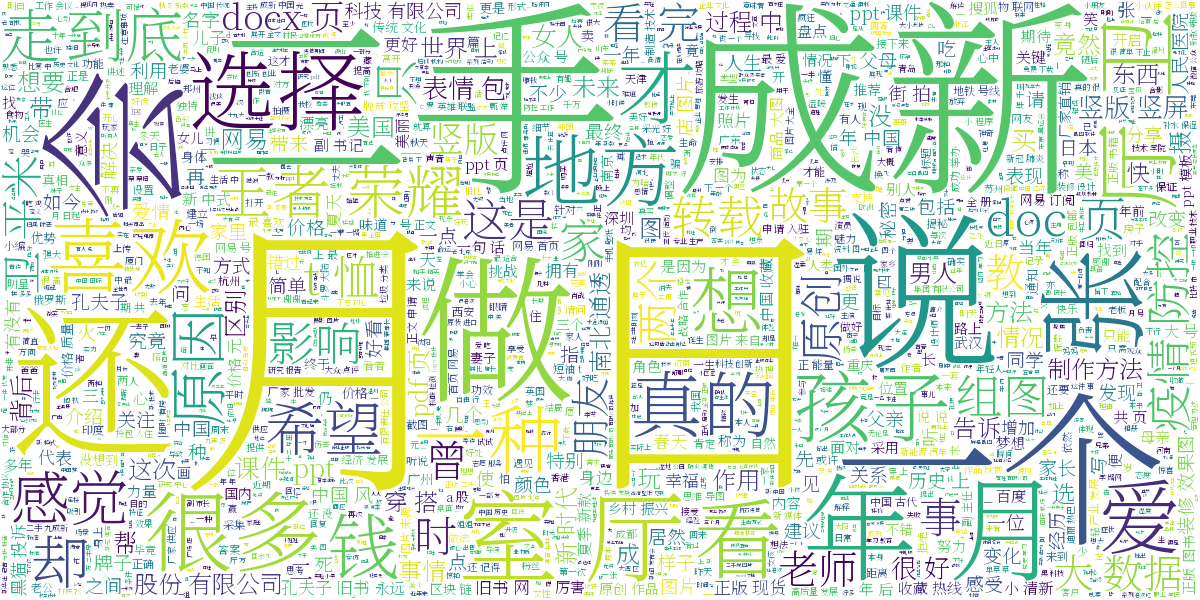}
  \caption{\label{fig:word-cloud} The word cloud generated with texts
    in Wukong dataset. For example, ``\texttt{月}'' means
    \textit{month}; ``\texttt{日}'' is \textit{day}; ``\texttt{做}''
    is \textit{do} and ``\texttt{一个}'' means \textit{one}.}
\end{figure}

\begin{figure}[!h]
  \centering
  \includegraphics[width=.9\linewidth]{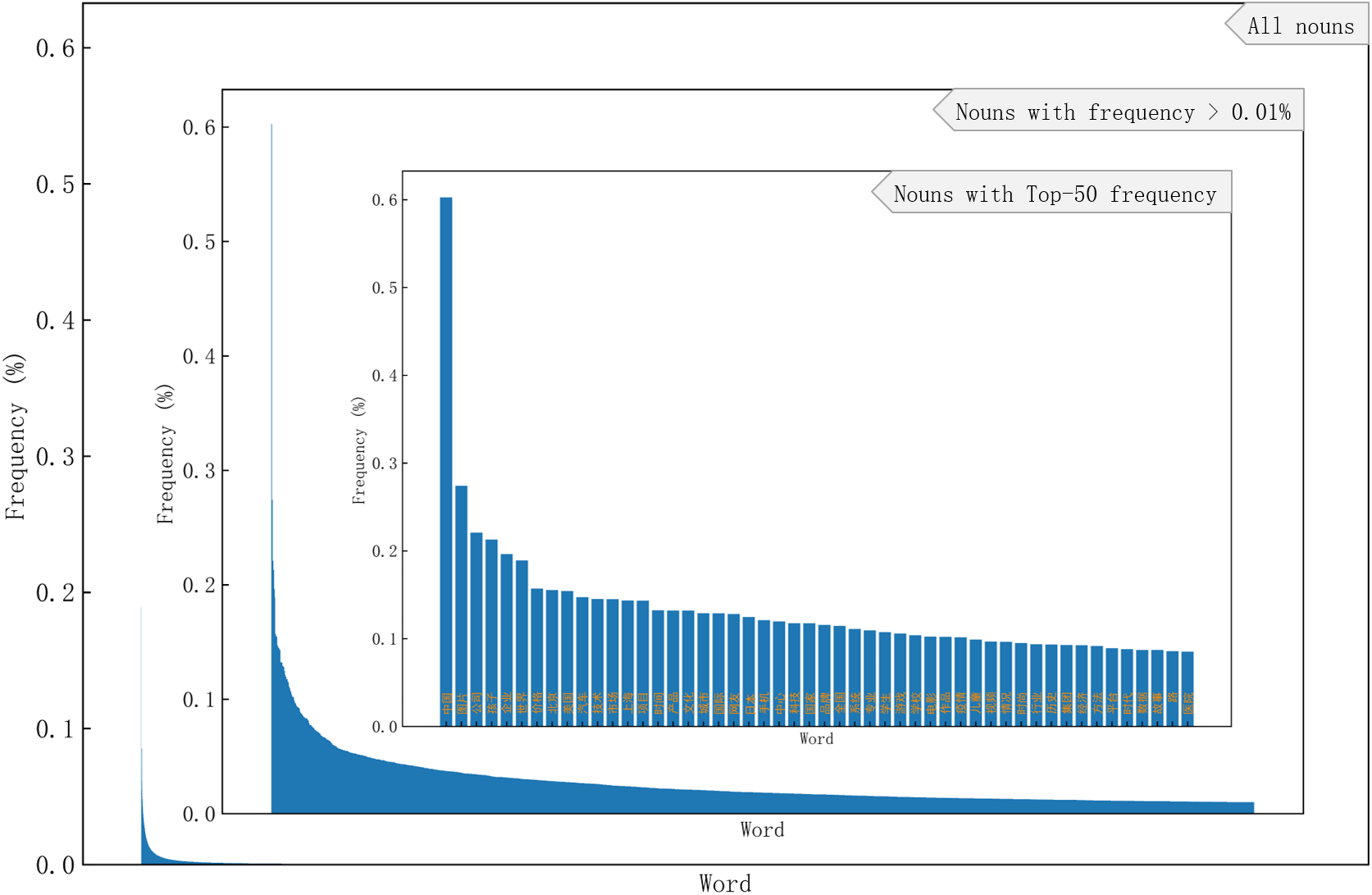}
  \caption{\label{fig:noun-barplot} The word frequency of nouns in our dataset. A wide range of concepts are covered.}
\end{figure}

Figure~\ref{fig:dataset_samples} shows some examples in our dataset. These image-text pairs involve many types of content, e.g., social news, sporting events, product introduction, et~al. Therefore, our dataset is suitable for general-purpose multi-modal pre-training. \added{Additionally, in Figure~\ref{fig:word-cloud}, we visualize the distribution of words (consisting of one or more tokens) in our dataset. We use the Chinese text segmentation module \textit{jieba}}\footnote{\url{https://github.com/fxsjy/jieba}} \added{to generate words and build this word cloud of our dataset.} \added{Additionally, for the topics or themes of the samples, Figure~\ref{fig:noun-barplot} shows the word frequency of nouns in our dataset. Naturally, a long tail distribution is followed and a wide range of concepts are covered.}

\section{Experimental Setup}

\begin{table}[h!]
  \setlength{\tabcolsep}{4pt}
  \centering
  \caption{\label{tab:model-arch} Detailed settings of our model
    variants. The resolution of image is $224 \times 224$ and 
    the length of text is 32.}
%   \footnotesize  % use small font size in this table
  \begin{tabular}{r|ccccccc}
    \toprule
    \multirow{2.5}{*}{Model} & \multirow{2.5}{*}{\shortstack{Image\\encoder}} & \multirow{2.5}{*}{\shortstack{Linear projected \\embeddings }} & \multirow{2.5}{*}{\shortstack{Token\\reduction}} & \multicolumn{3}{c}{Text encoder} & \multirow{2.5}{*}{\#Parameters} \\
    \cmidrule(lr){5-7}
    & & & & \#layers & \#heads & width & \\
    \midrule
    Wukong\textsubscript{ViT-B} & ViT-B/32  & 256 & 12 & 12 & 8 & 512 & \num{136}M \\
    Wukong\textsubscript{ViT-L} & ViT-L/14  & 256 & 24 & 12 & 12 & 768 & \num{404}M \\
    Wukong\textsubscript{Swin-L} & Swin-L & 256 & 12 & 12 & 12 & 768 & \num{297}M \\
    \bottomrule
  \end{tabular}
\end{table}

% \begin{table}[!h]
%   \centering
%   \caption{\label{tab:model-sizes} The number of parameters of different model variants.}
%   \begin{tabular}{r|ccc}
%     \toprule
%     \textbf{Model} & CLIP\textsubscript{ViT-B} & FILIP\textsubscript{ViT-B} & Wukong\textsubscript{ViT-B}\\
%     \textbf{\#Parameters} & \num{136774913} & \num{136447233} & \num{136467201}\\
%     \midrule
%     \textbf{Model} & CLIP\textsubscript{ViT-L} & FILIP\textsubscript{ViT-L} & Wukong\textsubscript{ViT-L}\\
%     \textbf{\#Parameters} & \num{405862913} & \num{404945409} & \num{404968449}\\
%     \midrule
%     \textbf{Model} & CLIP\textsubscript{Swin-L} & FILIP\textsubscript{Swin-L} & Wukong\textsubscript{Swin-L}\\
%     \textbf{\#Parameters} & \num{298408737} & \num{297228577} & \num{297248545}\\
%     \bottomrule
%   \end{tabular}
% \end{table}

\begin{table}[!h]
  \centering
  \caption{\label{tab:hyper-parameters} Hyper-parameters used in model training.}
%   \footnotesize  % use small font size in this table
  \begin{tabular}{ccccc}
    \toprule
    \multirow{2.5}{*}{\shortstack{Initial\\Temperature}} & \multicolumn{3}{c}{LAMB} & \multirow{2.5}{*}{\shortstack{Total\\Epochs}} \\
    \cmidrule(lr){2-4}
     & $\beta_1$ & $\beta_2$ & $\epsilon$ & \\
    \midrule
    0.07 & 0.9 & 0.999 & $10^{-2}$ & 20\\
    \bottomrule
  \end{tabular}
\end{table}

The experimental settings of our model variants are described in Table~\ref{tab:model-arch}.
For better generalization and data-efficiency, we employ
Autoaugment~\cite{cubuk2019autoaugment} for image data augmentation
that aims to build more image-text pairs. All of our models are
trained using Nvidia V100 GPUs and Ascend cards. Specifically,
Wukong\textsubscript{ViT-B} is trained using 32 GPUs for 3 days, Wukong\textsubscript{ViT-L} is trained using 32 GPUs for 10 days and
Wukong\textsubscript{Swin-L} is trained using 40 GPUs for 5
days. We use LAMB optimizer~\cite{2019Large} and cosine
learning rate schedule with a linear warmup~\cite{2016SGDR}. Weight
decay regularization is applied to all parameters except for bias,
layer normalization, token embedding, positional embedding and
temperature in the contrastive loss. The detailed hyper-parameters are
shown in Table~\ref{tab:hyper-parameters}. In order to pick the optimal
checkpoint out, ImageNet dataset~\cite{deng2009imagenet} with
translated class names is used for zero-shot validation.

\section{Supplementary Experiments}

\subsection{Tokenization for Chinese}

\begin{table}[!h]
  \centering
  \setlength{\tabcolsep}{4pt}
  \caption{\label{tab:char-word-token} Comparison of character-grained tokenization and word-grained tokenization method. The metric is top-1 accuracy (\%) of zero-shot image classification. The better result is highlighted with \textbf{bold}.}
%   \footnotesize  % use small font size in this table
  \begin{tabular}{l|*{10}{c}|{c}}
    \diagbox{Model}{Dataset} & \rotatebox[origin=c]{90}{CIFAR10} & \rotatebox[origin=c]{90}{CIFAR100} & \rotatebox[origin=c]{90}{Caltech101} & \rotatebox[origin=c]{90}{Caltech256} & \rotatebox[origin=c]{90}{DTD} & \rotatebox[origin=c]{90}{Sports} & \rotatebox[origin=c]{90}{Flowers} & \rotatebox[origin=c]{90}{SUN397} & \rotatebox[origin=c]{90}{EuroSAT} & \rotatebox[origin=c]{90}{ImageNet} & \rotatebox[origin=c]{90}{Average} \\
    \midrule
    Wukong\textsubscript{ViT-B-Word} & \textbf{89.1} & 62.1 & 88.7 & 80.8 & 29.1 & 93.7 & 53.3 & 49.6 & \textbf{36.2} & 43.9 & 62.65 \\
    Wukong\textsubscript{ViT-B} & 87.1 & \textbf{62.6} & \textbf{89.1} & \textbf{82.3} & \textbf{37.3} & \textbf{95.6} & \textbf{64.8} & \textbf{56.0} & 32.6 & \textbf{49.1} & \textbf{65.65} \\
    \bottomrule
  \end{tabular}
\end{table}

Table~\ref{tab:char-word-token} shows the comparison between using the character-grained and word-grained tokenization in our Wukong\textsubscript{ViT-B} model. We use the python module \textit{jieba} to do Chinese word segmentation to split Chinese text into words. All experimental settings remain the same expect for the tokenization. Results show that Wukong\textsubscript{ViT-B} achieve better performance than Wukong\textsubscript{ViT-B-Word}. We believe the main reason is that the character-grained tokens are more fine-grained than word-grained, since a Chinese word often contains more than one character. Such character-grained method contributes to help models learn the deep semantic token-wise similarity between an image patch with its paired fine-grained textual tokens. A typical example from the Chinese ImageNet dataset is that the word ``\texttt{蜂鸟}''(hummingbird) consists of two characters: ``\texttt{蜂}''(bee) and ``\texttt{鸟}'' (bird).

\subsection{Visualization of Word-patch Alignment}

Since we follow the fine-grained interaction in
FILIP~\cite{yao2021filip}, our trained models
% Wukong\rlap{\textsuperscript{F}}\textsubscript{ViT-L} 
FILIP\textsubscript{ViT-L} and FILIP\textsubscript{Swin-L}
% Wukong\rlap{\textsuperscript{F}}\textsubscript{Swin} 
likewise own the capability of capturing the correspondence between images and texts. 
% In this section, we use our pre-trained models
% Wukong\textsubscript{ViT-L} and Wukong\textsubscript{Swin}. 
Note that they are trained using the token-wise similarity. We exclude ones with the global similarity since they lack of word-patch alignment capability, which has been evidenced in previous work~\cite{yao2021filip}.
%Wukong\rlap{\textsuperscript{F}}\textsubscript{ViT-L} and Wukong\rlap{\textsuperscript{F}}\textsubscript{Swin} for visualization. 
%\footnote{\jiaxi{The models used for visualization need to be changed, for examples, Wukong\textsubscript{ViT-L} and Wukong\rlap{\textsubscript{ViT-L}}\textsuperscript{G}}}

\begin{figure}[ht!]
  \centering \includegraphics[width=.95\linewidth]{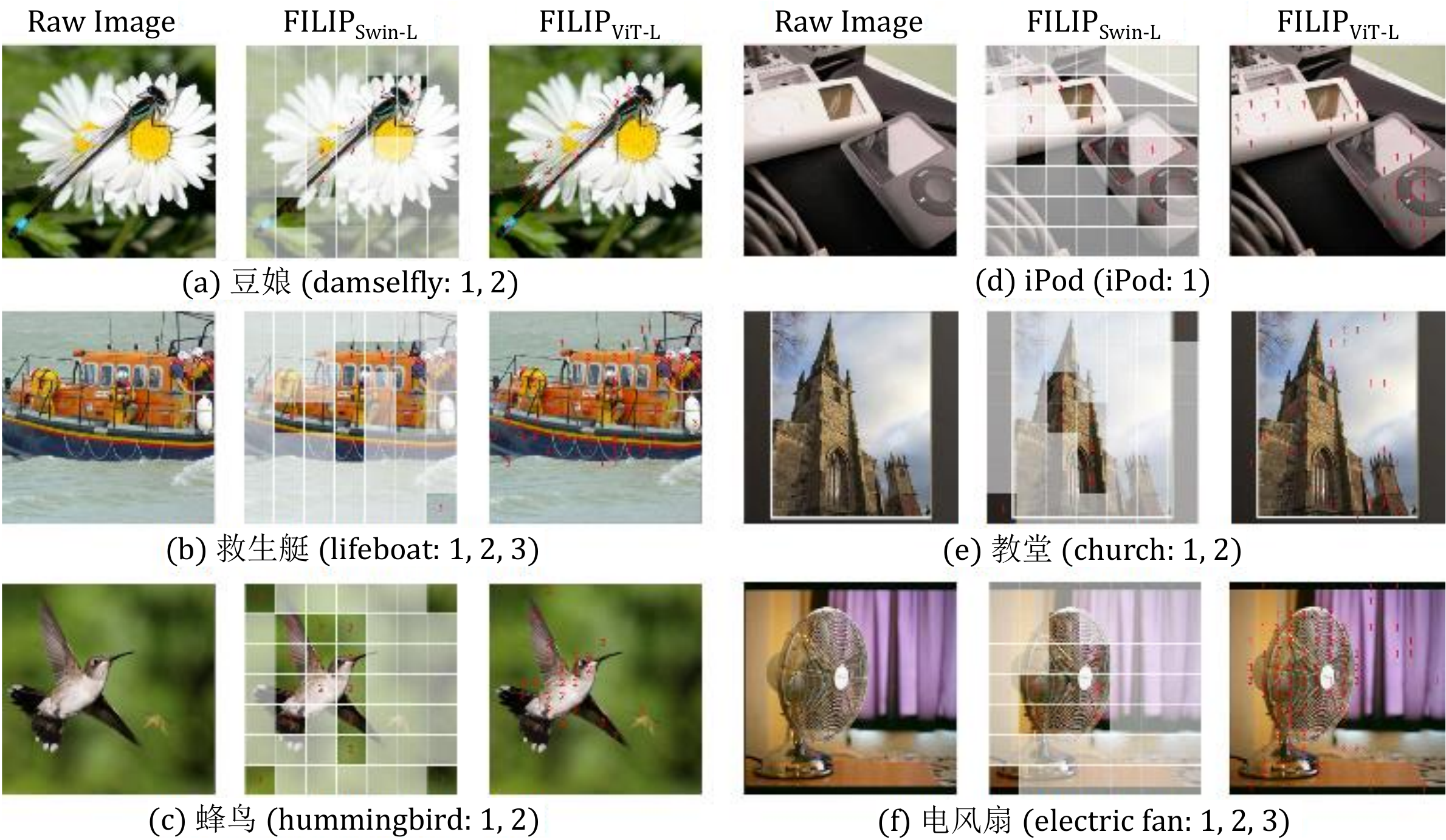}
\caption{Visualization of word-patch alignment. We randomly choose
    six classes in the Chinese ImageNet dataset. Each Chinese label
    name is used as a prompt, whose English text is described in the
    parentheses. Behind which, the tail numbers indicate the location
    indices of this class label in the tokenized textual input. Take
    (a) as an example, the number 0 always represents \texttt{[CLS]},
    the number 1 is the tokenized ``\texttt{豆}'' and the number 2 is
    ``\texttt{娘}''. Indices of
    the tokenized label name are highlighted in red.}
  \label{fig:vis}
\end{figure}

As shown in Figure~\ref{fig:vis}, we visualize images from six
labels in the Chinese ImageNet.
%(i.e., \textit{damselfly; lifeboat; hummingbird; iPod; church and electric fan}).
We apply the same visualization method as FILIP~\cite{yao2021filip}, to align textual tokens and image patch tokens. In particular, we calculate the token-wise similarity between each image patch token and all tokenized textual tokens from the text label, i.e., \texttt{[CLS]}\{class label tokens\}\texttt{[SEP]}. For each image patch, the position index of textual tokens with the maximum
similarity is considered as its predicted text token. Note that the
Chinese class label is often tokenized to more than one token. We
highlight all the predicted position indices that correspond to the
class label, and place them at the center of the corresponding patches. 
% Since we use the visual encoder ViT-L/14 in
% Wukong\textsubscript{ViT-L}, each image is patchified to 16$\times$16. For the used Swin-L Transformer in
% Wukong\textsubscript{Swin}, the output resolution is
% $\frac{H}{32}\times\frac{W}{32}$, that is, 7$\times$7
% patches. Therefore, Wukong\textsubscript{ViT-L} presents the more fine-cut grids than Wukong\textsubscript{Swin}.

From Figure~\ref{fig:vis}, we surprisingly find that both models are
able to predict image patches of the target object. For 
% Wukong\rlap{\textsuperscript{F}}\textsubscript{ViT-L} 
FILIP\textsubscript{ViT-L} with each image patchified to 16$\times$16, such word-patch
alignment is more fine-grained than FILIP\textsubscript{Swin-L}
% Wukong\rlap{\textsuperscript{F}}\textsubscript{Swin} 
with the output resolution as 7$\times$7. 
Take Figure~\ref{fig:vis}~(e) as an example, 
% Wukong\rlap{\textsuperscript{F}}\textsubscript{ViT-L} 
FILIP\textsubscript{ViT-L} is even able to align Chinese tokens ``\texttt{教}'' and ``\texttt{堂}'',
which means church as one word, to the smaller church in the
bottom-right corner. 
% Wukong\rlap{\textsuperscript{F}}\textsubscript{ViT-L} 
FILIP\textsubscript{ViT-L} also well outlines the
hummingbird in the example of Figure~\ref{fig:vis}~(c), while
% Wukong\rlap{\textsuperscript{F}}\textsubscript{Swin} 
FILIP\textsubscript{Swin-L} often aligns to the main body of the target object.
% However, since more fine-cut patches are presented, it might
% bring noises at some point compared to
% Wukong\textsubscript{Swin}. As in the (e) example, some obvious wrongly predicted patches can be viewed for Wukong\textsubscript{ViT},
% and similarly in Figure~\ref{fig:vis}~(f), some image patches surrounding the fan are predicted to token index 1. Note that this token ``\texttt{电}'' of
% index 1 means electricity, which essentially is not direct to the
% meaning of fan. 
Another interesting observation is that these Chinese pre-trained models are able to align image patches to English tokens as shown in Figure~\ref{fig:vis}~(d). The main reason lies in that the vocabulary used from BERT~\cite{devlin2018bert} also includes multilingual words such as ``\texttt{iPod}''.

Overall, this visualization confirms that our released models pre-trained on Wukong dataset indeed learn the correspondence between images and Chinese texts, or even in a more finer-grained manner, the alignment between image patches and words. This capability of aligning words and patches offers a potential solution for image object localization.

% This visualization of word-patch alignment evidences the effectiveness
% of cross-modal token-wise similarity even in the LiT-tuning setting.
% Though the visual encoder (i.e., ViT-L or Swin-L) is locked in the
% pre-training phase, the learnable linear projection layer on top of
% it, is still able to align patches and words in a fine-grained
% manner. We also find that this token-wise similarity in loss function
% works across various patch-based visual encoders, not only the
% traditional ViT architecture but also SwinT that uses the
% sophisticated shifted windowing attention and patch merging. Given
% this promising capability of aligning words and patches, our
% experiment offers a potential solution towards image object
% localization.

\section{Downstream Datasets}

\subsection{Prompt Template}
\label{appendix_prompt}

As previously observed in GPT-3~\cite{brown2020language}, the
zero-shot performance can be significantly improved by customizing the
prompt templates to each task. CLIP~\cite{radford2021learning} also
shows that specifying the category for each dataset contributes to the
performance. However, since we only aim to provide a Chinese dataset
with a general benchmarking of our released models, we leave the
``prompt engineering'' to the future work. We simply use the reported
80 general English prompts in CLIP and translate them to Chinese
manually, as follows. Note that ``\{\}'' is replaced by the exact
Chinese label name. We release these Chinese prompts for future fair comparison in the community. \added{Below are all the 80 Chinese prompts and the corresponding English prompts.}

\paragraph{Chinese Prompts:} ``\{\}的照片。'', ``许多\{\}的照片。'',
``一张包含\{\}的照片。'', ``质量差的\{\}的照片。'', ``\{\}的雕塑。'',
``难以看到\{\}的照片。'', ``\{\}的低分辨率照片。'', ``\{\}的渲染。'',
``涂鸦\{\}。'', ``\{\}的糟糕照片。'', ``\{\}的裁剪照片。'', ``\{\}的纹
身。'', ``\{\}的刺绣照片。'', ``很难看到\{\}的照片。'', ``\{\}的明亮照
片。'', ``一张干净的\{\}的照片。'', ``\{\}的深色照片。'', ``\{\}的手绘
画。'', ``我的\{\}的照片。'', ``不自然的\{\}的照片。'', ``一张酷
的\{\}的照片。'', ``\{\}的特写照片。'', ``\{\}的黑白照片。'',
``一幅\{\}的画。'', ``一幅\{\}的绘画。'', ``一张\{\}的像素照片。'',
``\{\}的雕像。'', ``一张\{\}的明亮照片。'', ``\{\}的裁剪照片。'', ``人
造的\{\}的照片。'', ``一张关于\{\}的照片。'', ``损坏的\{\}的jpeg照
片。'', ``\{\}的模糊照片。'', ``\{\}的相片。'', ``一张\{\}的好照片。'',
``\{\}的渲染照。'', ``视频游戏中的\{\}。'', ``一张\{\}的照片。'',
``\{\}的涂鸦。'', ``\{\}的近距离照片。'', ``\{\}的折纸。'', ``\{\}在视
频游戏中。'', ``\{\}的草图。'', ``\{\}的涂鸦照。'', ``\{\}的折纸形
状。'', ``低分辨率的\{\}的照片。'', ``玩具\{\}。'', ``\{\}的副本。'',
``\{\}的干净的照片。'', ``一张大\{\}的照片。'', ``\{\}的重现。'', ``一
张漂亮的\{\}的照片。'', ``一张奇怪的\{\}的照片。'', ``模糊的\{\}的照
片。'', ``卡通\{\}。'', ``\{\}的艺术作品。'', ``\{\}的素描。'',
``刺绣\{\}。'', ``\{\}的像素照。'', ``\{\}的拍照。'', ``\{\}的损坏的照
片。'', ``高质量的\{\}的照片。'', ``毛绒玩具\{\}。'', ``漂亮的\{\}的照
片。'', ``小\{\}的照片。'', ``照片是奇怪的\{\}。'', ``漫画\{\}。'',
``\{\}的艺术照。'', ``\{\}的图形。'', ``大\{\}的照片。'', ``黑白
的\{\}的照片。'', ``\{\}毛绒玩具。'', ``一张\{\}的深色照片。'',
``\{\}的摄影图。'', ``\{\}的涂鸦照。'', ``玩具形状的\{\}。'',
``拍了\{\}的照片。'', ``酷酷的\{\}的照片。'', ``照片里的小\{\}。'',
``\{\}的刺青。''

\paragraph{English Prompts:} \added{``a photo of a \{\}.'', ``a bad photo of a \{\}.'', ``a photo of many \{\}.'', ``a sculpture of a \{\}.'', ``a photo of the hard to see \{\}.'', ``a low resolution photo of the \{\}.'', ``a rendering of a \{\}.'', ``graffiti of a \{\}.'', ``a bad photo of the \{\}.'', ``a cropped photo of the \{\}.'', ``a tattoo of a \{\}.'', ``the embroidered \{\}.'', ``a photo of a hard to see \{\}.'', ``a bright photo of a \{\}.'', ``a photo of a clean \{\}.'', ``a photo of a dirty \{\}.'', ``a dark photo of the \{\}.'', ``a drawing of a \{\}.'', ``a photo of my \{\}.'', ``the plastic \{\}.'', ``a photo of the cool \{\}.'', ``a close-up photo of a \{\}.'', ``a black and white photo of the \{\}.'', ``a painting of the \{\}.'', ``a painting of a \{\}.'', ``a pixelated photo of the \{\}.'', ``a sculpture of the \{\}.'', ``a bright photo of the \{\}.'', ``a cropped photo of a \{\}.'', ``a plastic \{\}.'', ``a photo of the dirty \{\}.'', ``a jpeg corrupted photo of a \{\}.'', ``a blurry photo of the \{\}.'', ``a photo of the \{\}.'', ``a good photo of the \{\}.'', ``a rendering of the \{\}.'', ``a \{\} in a video game.'', ``a photo of one \{\}.'', ``a doodle of a \{\}.'', ``a close-up photo of the \{\}.'', ``the origami \{\}.'', ``the \{\} in a video game.'', ``a sketch of a \{\}.'', ``a doodle of the \{\}.'', ``a origami \{\}.'', ``a low resolution photo of a \{\}.'', ``the toy \{\}.'', ``a rendition of the \{\}.'', ``a photo of the clean \{\}.'', ``a photo of a large \{\}.'', ``a rendition of a \{\}.'', ``a photo of a nice \{\}.'', ``a photo of a weird \{\}.'', ``a blurry photo of a \{\}.'', ``a cartoon \{\}.'', ``art of a \{\}.'', ``a sketch of the \{\}.'', ``a embroidered \{\}.'', ``a pixelated photo of a \{\}.'', ``itap of the \{\}.'', ``a jpeg corrupted photo of the \{\}.'', ``a good photo of a \{\}.'', ``a plushie \{\}.'', ``a photo of the nice \{\}.'', ``a photo of the small \{\}.'', ``a photo of the weird \{\}.'', ``the cartoon \{\}.'', ``art of the \{\}.'', ``a drawing of the \{\}.'', ``a photo of the large \{\}.'', ``a black and white photo of a \{\}.'', ``the plushie \{\}.'', ``a dark photo of a \{\}.'', ``itap of a \{\}.'', ``graffiti of the \{\}.'', ``a toy \{\}.'', ``itap of my \{\}.'', ``a photo of a cool \{\}.'', ``a photo of a small \{\}.'', ``a tattoo of the \{\}.''}

\subsection{Datasets for Image-text Retrieval}
\label{appendix_dataset}

The data scale of datasets for image-text retrieval is described in Table~\ref{tab:retrieval-dataset}. The texts in Flickr8K-CN,
COCO-CN, AIC-ICC are human-annotated, the texts in Flickr30K-CN
train/val set are machine-translated while the texts in Flickr30K-CN
test set are human-translated from their original English
counterparts. In Flickr8K-CN, Flickr30K-CN and
AIC-ICC, each image is paired with 5 texts. In COCO-CN, each image
is paired with 1 to 2 texts. In MUGE, each
text is paired with 1 to 2 images in the train set, and with about 6 images in the val/test sets.

\begin{table}[ht!]
  \centering
  \caption{\label{tab:retrieval-dataset} Statistics of each image-text
    retrieval dataset.}
  \begin{tabular}{clrr}
    \toprule
    \textbf{Dataset} & \textbf{split} & \textbf{\#Images} & \textbf{\#Sentences} \\
    \midrule
    \multirow{3}{*}{Flickr8K-CN~\cite{flickr8k-cn}} & train & \num{6000} & \num{30000} \\
            & val & \num{1000} & \num{5000} \\
            & test & \num{1000} & \num{5000} \\
    \midrule[.2pt]
    \multirow{3}{*}{Flickr30K-CN~\cite{flickr30k-cn}} & train & \num{29783} & \num{148915} \\
            & val & \num{1000} & \num{5000} \\
            & test & \num{1000} & \num{5000} \\
    \midrule[.2pt]
    \multirow{3}{*}{COCO-CN~\cite{coco-cn}} & train & \num{18341} & \num{20065} \\
            & val & \num{1000} & \num{1100} \\
            & test & \num{1000} & \num{1053} \\
    \midrule[.2pt]
    \multirow{4}{*}{AIC-ICC~\cite{aicicc}} & train & \num{210000} & \num{1050000} \\
            & val & \num{30000} & \num{150000} \\
            & test-1 & \num{30000} & \num{150000} \\
            & test-2 & \num{30000} & \num{150000} \\
    \midrule[.2pt]
    \multirow{3}{*}{MUGE~\cite{lin2021m6}} & train & \num{129380} & \num{248786} \\
            & val & \num{29806} & \num{5008} \\
            & test & \num{30399} & \num{5004} \\
    \midrule[.2pt]
    Wukong-Test & val & \num{33365} & \num{33365} \\
    \bottomrule
 \end{tabular}
\end{table}

\section{Limitations and Societal Impacts}

% limitation:
% 1. 语言随着人类活动而发展，图文数据集仅能表达当前概念，无法覆盖未来可能出现的概念或术语。
% 2. 由于数据集来自于中文互联网，语言的词汇和表述可能存在bias，可能存在书面语多于口语的情况、语言习惯相对出版物可能存在差异。
% 3. 出于数据集使用和发布的考虑，数据集中不包含特长句，因此模型对于长文或复杂概念的理解能力有限。
% societal impacts
% 1. 该数据集规模大且不限定领域，在使用中可能存在对性别、种族、地域的偏见，在面向用户的应用上可能会产生undesirable的结果。

Wukong dataset might only contain the current concepts and language expression at the time of collection.
Since language evolves with human activities, our dataset certainly cannot cover the newly emerging concepts, words and language expression in the future. It is the same case for the image data side, where the new visual object or design can not be covered. However, fine-tuning pre-trained models on these up-to-date data is able to address this issue.
In addition, our dataset is built on the corpora from Chinese Internet, which means the vocabulary and expression may more or less fit into the Chinese culture. Also, there is more written language than spoken language and it might bring bias at some point.
Another limitation is the absence of very long texts in our dataset. Therefore, the ability of understanding documents using our released models might be limited. Furthermore, in terms of societal impacts, our dataset is built in a general purpose with images and texts collected from unlimited domains. Models trained on this dataset might express some undesirable and uncontrollable tendencies in terms of image-text correspondence. 
% of discrimination or even hatred, e.g., age, gender, race, et~al. 
Therefore, although our released models are discriminative, special attention is still suggested in practical use.
% especially for the customer-facing applications.

\section{Hosting and Maintenance Plan}

\added{Long-term maintenance of Wukong, as well as Wukong-Test, and models proposed and evaluated in our paper will be made
by the authors. The dataset website containing introductions,
benchmarks, terms of use and any possible improvement in the future
are hosted in Github Pages which is a widely-used website hosting
service. In terms of content hosting, there are three parts: code,
models and datasets. All of them are hosted on open platforms
that each individual is able to download freely. For
evaluation code, Pytorch version is hosted on Github and Mindspore
version is hosted on Gitee, an open-source code hosting platform
specialized for Chinese users. The model checkpoints trained in our
paper are hosted on Google Drive. The datasets including Wukong and
Wukong-Test are hosted on Google Drive and Baidu Cloud, a
widely-used cloud storage service in China, as backup.}

\section{License}

Unless specifically labeled otherwise, our released datasets are provided to You under the terms of the Creative Commons Attribution-NonCommercial-ShareAlike 4.0 International Public License (“CC BY-NC-SA 4.0”), with the additional terms included herein. The CC BY-NC-SA 4.0 may be accessed at \url{https://creativecommons.org/licenses/by-nc-sa/4.0/legalcode}. When You download or use the datasets from our website or elsewhere, You are agreeing to comply with the terms of CC BY-NC-SA 4.0, and also agreeing to the dataset Terms. Where these dataset Terms conflict with the terms of CC BY-NC-SA 4.0, these dataset Terms shall prevail. We reiterate once again that this dataset is used only for non-commercial purposes such as academic research, teaching, or scientific publications. We prohibits You from using the dataset or any derivative works for commercial purposes, such as selling data or using it for commercial gain.

\end{CJK*}
\end{document}